\documentclass[nohyperref]{article}

\usepackage{microtype}
\usepackage{graphicx}
\usepackage{subfigure}
\usepackage{booktabs} 

\usepackage{hyperref}

\usepackage[accepted]{icml2022}

\usepackage{amsmath}
\usepackage{amssymb}
\usepackage{mathtools}
\usepackage{amsthm}

\usepackage[capitalize,noabbrev]{cleveref}

\usepackage{booktabs}
\usepackage{multirow}

\usepackage{xcolor,pifont}
\newcommand{\cmark}{\ding{52}}
\newcommand{\xmark}{\ding{56}}

\newcommand{\ie}{i.e.}

\theoremstyle{plain}

\theoremstyle{definition}

\theoremstyle{remark}

\usepackage[textsize=tiny]{todonotes}

\icmltitlerunning{AutoSNN: Towards Energy-Efficient Spiking Neural Networks}

\begin{document}

\twocolumn[
\icmltitle{AutoSNN: Towards Energy-Efficient Spiking Neural Networks}



\icmlsetsymbol{equal}{*}
\icmlsetsymbol{student}{$\dagger$}

\begin{icmlauthorlist}
\icmlauthor{Byunggook Na}{sait,student}
\icmlauthor{Jisoo Mok}{snu} 
\icmlauthor{Seongsik Park}{kist} 
\icmlauthor{Dongjin Lee}{snu} 
\icmlauthor{Hyeokjun Choe}{snu} 
\icmlauthor{Sungroh Yoon}{snu,snu_others}
\end{icmlauthorlist}

\icmlaffiliation{sait}{Samsung Advanced Institute of Technology, South Korea}
\icmlaffiliation{snu}{Department of Electric and Computer Engineering, Seoul National University, South Korea}
\icmlaffiliation{kist}{Korea Institute of Science and Technology, South Korea}
\icmlaffiliation{snu_others}{Interdisciplinary Program in Artificial Intelligence, Seoul National University, South Korea}

\icmlcorrespondingauthor{Sungroh Yoon}{sryoon@snu.ac.kr}

\icmlkeywords{Spiking Neural Networks, Neural Architecture Search}

\vskip 0.3in
]



\printAffiliationsAndNotice{$^\dagger$Portions of this research were done while the author was a Ph.D. student in SNU.}  

\begin{abstract}
Spiking neural networks (SNNs) that mimic information transmission in the brain can energy-efficiently process spatio-temporal information through discrete and sparse spikes, thereby receiving considerable attention.
To improve accuracy and energy efficiency of SNNs, most previous studies have focused solely on training methods, and the effect of architecture has rarely been studied.
We investigate the design choices used in the previous studies in terms of the accuracy and number of spikes and figure out that they are not best-suited for SNNs.
To further improve the accuracy and reduce the spikes generated by SNNs, we propose a spike-aware neural architecture search framework called AutoSNN.
We define a search space consisting of architectures without undesirable design choices.
To enable the spike-aware architecture search, we introduce a fitness that considers both the accuracy and number of spikes.
AutoSNN successfully searches for SNN architectures that outperform hand-crafted SNNs in accuracy and energy efficiency.
We thoroughly demonstrate the effectiveness of AutoSNN on various datasets including neuromorphic datasets. 

\end{abstract}

\section{Introduction}\label{sec:intro}


Spiking neural networks (SNNs) are the next generation of neural networks inspired by the brain's information processing systems~\cite{maass1997networks}.
The neurons in SNNs asynchronously transmit information through sparse and binary spikes, enabling event-driven computing.
Unlike conventional neural networks being executed on GPUs, energy consumption occurs if a spike is generated on neuromorphic chips, which support neuromorphic computing.
Hence, SNNs can significantly improve the energy efficiency of artificial intelligence systems.
Most neuromorphic chips adopt network-on-chip architectures with neuromorphic cores, and SNNs are mapped to these multiple cores~\cite{davies2018loihi,merolla2014million}.
A large number of spikes cause spike congestion between the cores, thereby considerably increasing the communication overhead and energy consumption~\cite{davies2021advancing}.
Therefore, when realizing SNNs on neuromorphic chips, their energy efficiency, that is, the number of generated spikes, must be considered along with accuracy.

\begin{figure}[t]
    \centering
    \includegraphics[width=\linewidth]{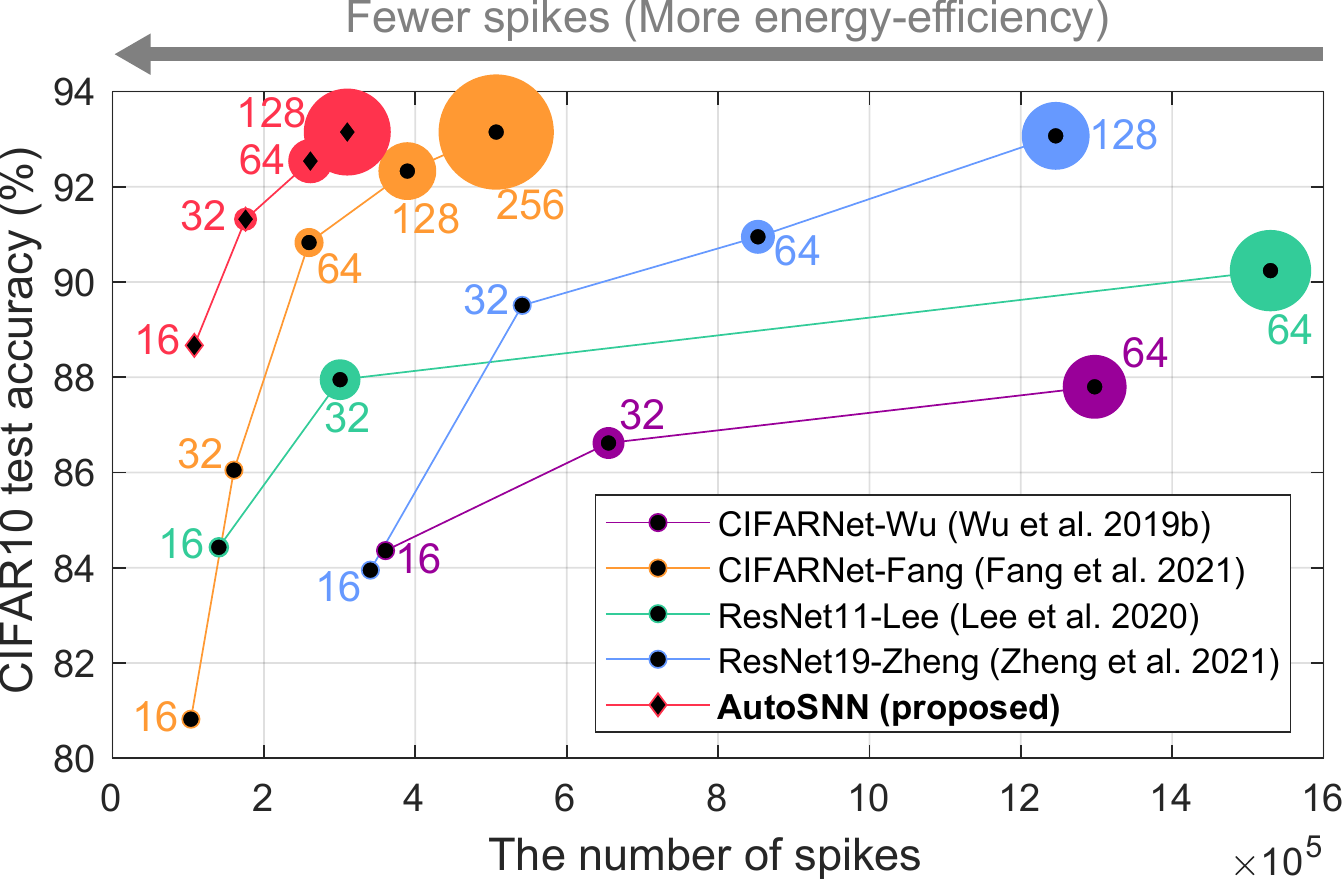}
    \vspace{-1em}
    \caption{The number of spikes and CIFAR10 accuracy of various SNN architectures. Black circle- and diamond-shaped markers denote hand-crafted and automatically-designed SNN architectures, respectively. The size of a colored circle is proportionate to the model size and the number next to the circle indicates the initial channel of each SNN architecture. \textbf{AutoSNN discovers an energy-efficient SNN architecture that outperforms hand-crafted SNNs in terms of the accuracy and number of spikes.}}
    \label{fig:spikes_vs_acc}
\end{figure}

As a means of improving the performance of SNNs and relieving their energy consumption, previous studies only focused on training algorithms and protocols, such as reducing the timesteps required in SNNs~\cite{neftci2019surrogate,wu2019direct,fang2021incorporating,lee2020enabling,he2020comparing,kaiser2020synaptic,zheng2021going}.
They employed conventional architectures typically used as artificial neural networks (ANNs), such as VGGNet~\cite{Simonyan15vggnet} and ResNet~\cite{he2016resnet}.
Even though architecture modifications such as spike-element-wise residual blocks have been proposed~\cite{kim2020BNTT,kim2021SALT,fang2021SEW}, the architectural consideration about which architecture is suitable for use as SNNs in terms of the number of generated spikes has been overlooked.
As observed in \figurename~\ref{fig:spikes_vs_acc}, the number of generated spikes differs significantly depending on the SNN architecture.
It is necessary to investigate the design choices that affect the accuracy and spike generation.

In this study, we analyze architectural properties in terms of the accuracy and number of spikes, and identify preferable design choices for energy-efficient SNNs with minimal spikes.
The use of the global average pooling layer~\cite{lin2013nin_gap} and employing layers with trainable parameters for down-sampling~\cite{he2016resnet,sandler2018mobilenetv2,yang2019darts} decreases the energy efficiency of SNNs, suggesting the exclusion of these design choices from architectures.
To further improve the performance and energy efficiency, we adopt neural architecture search (NAS) that has emerged as an attractive alternative to hand-crafting ANN architectures.
Across various applications~\cite{xie2020survey,chen2019detnas,jiang2020spnas,guo2020hitdetector,kim2020est,ding2021hrnas,yan2021lighttrack,zhang2021dcnas}, NAS successfully searched for ANN architectures best-suited for the target objectives.
Inspired by the success of NAS, we propose a spike-aware NAS framework, named \textit{AutoSNN}, to design energy-efficient SNNs.

For AutoSNN, we define a search space considering both accuracy and energy efficiency and propose a spike-aware search algorithm.
To construct an expressive search space, we introduce a two-level search space that consists of a macro-level backbone architecture and micro-level candidate spiking blocks.
To explore the proposed search space with a reasonable search cost, we exploit the one-shot weight-sharing approach of the NAS~\cite{bender2018understanding,cai2020ofanet,guo2020spos,li2020random,zhang2021rlnas,yan2021lighttrack}.
To estimate the accuracy and number of spikes of candidate SNN architectures, a super-network that encodes all the architectures is trained through a direct training method for SNNs.
Once the super-network is trained, AutoSNN executes an evolutionary search algorithm that is proposed to find the SNN with the highest evaluation metric, which we call fitness.
To enable a spike-aware search, we define new fitness that reflects the accuracy and number of spikes.

AutoSNN discovers desirable SNN architectures that outperform hand-crafted SNNs in terms of both the accuracy and number of spikes, as shown in \figurename~\ref{fig:spikes_vs_acc}.
The superiority of the searched SNNs is consistently observed across various datasets including neuromorphic datasets.
Additionally, when our search algorithm is executed on a search space consisting of ANNs that share the proposed macro architecture, the resulting architectures with spiking blocks experience performance deterioration, emphasizing the importance of searching in the SNN search space to consider properties of SNNs.
The code of AutoSNN is available at https://github.com/nabk89/AutoSNN.
Our contributions are summarized as follows: 
\begin{itemize}
    \item To the best our knowledge, this is the first study to thoroughly investigate the effect of architectural components on performance and energy efficiency of SNNs.
    \item We propose a spike-aware NAS framework, named AutoSNN, and discover SNNs that outperform hand-crafted SNNs designed without consideration of the structures suitable for SNNs.
    \item We demonstrate the effectiveness of AutoSNN through substantial results and evaluations on various datasets.
\end{itemize}

\section{Background and Related Work}\label{sec:related_work}

\subsection{Spiking Neural Networks}

A spiking neuron in SNNs integrates synaptic inputs from the previous layer into an internal state called the membrane potential.
When the membrane potential integrated over time exceeds a certain threshold value, the neuron fires a spike to the next layer, and thus the spiking neurons transmit information through binary spike trains.
Among several spiking neuron models, the leaky integrate-and-fire (LIF) neuron is simple yet widely used due to its effectiveness~\cite{gerstner2002spiking}.
The dynamics of the LIF neuron at timestep $t$ are as follows:
\begin{equation}
\label{eq:LIF_dynamics}
    \tau_\mathrm{decay} \frac{\partial{V_\mathrm{mem}(t)}}{\partial{t}} = -(V_\mathrm{mem}(t) - V_\mathrm{reset}) + z(t),
\end{equation}
where $V_\mathrm{mem}$ is the membrane potential of a neuron and $V_\mathrm{reset}$ is the value of $V_\mathrm{mem}$ after spike firing.
$\tau_\mathrm{decay}$ is a membrane time constant that controls the decay of $V_\mathrm{mem}$ and $z$ is the presynaptic inputs.
For effective simulations with GPUs, Eq.~\ref{eq:LIF_dynamics} is re-written with an iterative and discrete-time form as follows:
\begin{equation}\label{eq:LIF_dynamics_discrete}
\begin{split}
    z^l [t] = & \sum_i w_i^l \phi_i^{l-1} [t] + b^l, \\
    H^l [t] = & V_\mathrm{mem}^l [t-1] + \\
    & \frac{1}{\tau_\mathrm{decay}}(-(V_\mathrm{mem}^l [t-1] - V_\mathrm{reset}) + z^l [t]), \\
    \phi^l [t] = & \Theta(H^l [t] - V_\mathrm{th}), \\
    V_\mathrm{mem}^l [t] = & H^l [t](1-\phi^l [t]) + V_\mathrm{reset}\phi^l [t],
\end{split}
\end{equation}
where superscript $l$ indicates the layer index.
The value of membrane potential at timestep $t$ is divided into two states, which represent the values before and after the trigger of a spike, denoted by $H[t]$ and $V_\mathrm{mem}[t]$, respectively.
$\phi[t]$ is a binary spike at $t$, $w$ is the synaptic weight, and $b$ is the bias.
The value of $\phi[t]$ is determined using $\Theta(x)$, the Heaviside step function that outputs 1 if $x \geq 0$ or 0 otherwise, and $V_\mathrm{th}$ is the threshold voltage for firing.
In this study, we employed parametric LIF (PLIF) neurons~\cite{fang2021incorporating} that was proposed to improve the performance of SNNs.
In PLIF neurons, $\frac{1}{\tau_\mathrm{decay}}$ is replaced with a sigmoid function $1 / (1+exp(-\alpha))$ with a trainable parameter $\alpha$.

To obtain a high-performing SNN, researchers have proposed various training methods that can be clustered into several approaches.
In the ANN-to-SNN conversion approach~\cite{diehl2015fast,rueckauer2017conversion,sengupta2019going,park2019burst,park2020t2fsnn,han2020rmpsnn,kim2020spiking}, after training an ANN, the optimized weights are transferred to the corresponding SNN, such that the firing patterns of the spiking neurons are encoded to approximate the activation values of the ANNs.
Even though these converted SNNs achieve accuracies comparable to those of ANNs, they heavily rely on the performance of trained ANNs and require a significant number of timesteps, which leads to significant spike generation and energy inefficiency.

As an approach to directly optimize SNNs, unsupervised learning methods based on the spike-timing-dependent plasticity method~\cite{diehl2015unsupervised} were introduced but were restricted to shallow networks and yielded limited performance.
Another approach is supervised learning based on a backpropagation algorithm~\cite{bohte2002error}.
A surrogate gradient function was used for backpropagation to approximate the gradients in the non-differentiable spiking activities~\cite{neftci2019surrogate,wu2019direct,lee2020enabling,he2020comparing,kaiser2020synaptic}; the detailed explanation is provided in Section~\ref{supp:direct_training}.
In recent studies, supervised learning has been effective for deep SNNs and yielded high accuracies with few timesteps and sparse generation of spikes~\cite{fang2021incorporating,zheng2021going}.
Therefore, we adopt a supervised learning method~\cite{fang2021incorporating} to obtain energy-efficient SNNs.

\subsection{Neural Architecture Search}

Early NAS methods~\cite{real2017large,real2019regularized,le2017naswithRL,zoph2018nasnet} sampled and separately evaluated candidate architectures, all of which had to be trained from scratch until convergence.
To reduce the enormous search cost induced by such approach, recent NAS methods have adopted the weight-sharing strategy~\cite{dean2018enas}.
With weight-sharing, search space $\mathcal{A}$, a set of all candidate architectures, is encoded in the form of a super-network $\mathcal{S}(W)$, whose weights $W$ are shared across all subnetworks.

Depending on how weight-sharing is incorporated into the search algorithm, NAS methods are primarily categorized into differentiable and one-shot methods.
The family of differentiable NAS methods~\cite{cai2019proxylessnas,yang2019darts,wu2019fbnet,dong2019gdas} starts by constructing a continuous search space that spans the entire search space of discrete architectures.
They introduced trainable architecture parameters $a$ into a super-network.
During the training of the super-network, $W$ and $a$ are optimized alternately, and once the training process is complete, an architecture with the largest $a$ is selected.

Unlike differentiable NAS, one-shot weight-sharing NAS~\cite{bender2018understanding,cai2020ofanet,you2020greedynas,guo2020spos,zhang2020oneshot_novelty,li2020random,peng2020cream,zhang2021rlnas,yan2021lighttrack} disentangles the search process into two procedures: super-network training and architecture evaluation.
During super-network training, architectures are sampled, and thus all candidate architectures in $\mathcal{A}$ can be approximately trained.
Given the trained super-network $\mathcal{S}(W^\ast)$, architectures inherit weights from $\mathcal{S}(W^\ast)$ and are evaluated without additional training.
Various sampling strategies for super-network training have been proposed to improve the reliability of the evaluation process, such as greedy path filtering~\cite{you2020greedynas}, novelty-based sampling~\cite{zhang2020oneshot_novelty}, and prioritized path distillation~\cite{peng2020cream}.
Because one-shot weight-sharing NAS does not have to retrain the super-network every time it needs to search for a new architecture~\cite{cai2020ofanet}, it is far more computationally efficient than differentiable NAS when deploying a diverse set of SNNs for different neuromorphic chips.

SNASNet~\cite{kim2022SNASNet}, which is concurrent to AutoSNN, has been recently proposed to find a performative SNN using a NAS method without training.
SNASNet only focuses on the accuracy of SNNs, while AutoSNN considers both their accuracy and energy efficiency.


\section{Architectural Analysis for SNNs}\label{sec:SNN_analysis}

The SNN architectures used in previous studies~\cite{neftci2019surrogate,wu2019direct,lee2020enabling,he2020comparing,kaiser2020synaptic,fang2021incorporating,zheng2021going} originated from conventional ANNs, such as VGGNet-styled stacked convolutional layers and max pooling layers~\cite{Simonyan15vggnet} and ResNet-styled stacked residual blocks with skip connections~\cite{he2016resnet}.
These architectures were selected without considering the architectural suitability of the SNNs.
This section analyzes architectural factors that affect the accuracy and spike generation of SNNs and investigates which design choices are desirable for energy-efficient SNNs.

We start by standardizing the building blocks used in previous SNN architectures~\cite{wu2019direct,lee2020enabling,fang2021incorporating,zheng2021going} into spiking convolution block (SCB) and spiking residual block (SRB), which stem from VGGNet and ResNet, respectively.
As depicted in \figurename~\ref{fig:spiking_blocks}, both SCB and SRB consist of two convolutional layers with spiking neurons, and the SRB additionally includes a skip connection.
In the SNN research field, spiking blocks with a kernel size of 3 have been widely used.

Using these spiking blocks, we discuss two architectural aspects: 1) A global average pooling (GAP) layer with spiking neurons is not suitable for SNNs, and 2) max pooling layers are best-suited for down-sampling in SNNs.
We prepare four architectures, denoted by SNN\_$\{1, 2, 3, 4\}$ and depicted in \figurename~\ref{supp_fig:macro_vs_variants}, consisting of to-be-determined (TBD) blocks that can be filled with spiking blocks.
Based on the SNN architecture~\cite{fang2021incorporating}, which yields the Pareto frontier curve in \figurename~\ref{fig:spikes_vs_acc} among previous studies, we construct SNN\_1 and its variants, described in the following sections.
In these architectures, normal blocks preserve the spatial resolution of the input feature map, and down-sampling (DS) blocks halve the spatial resolution.
We employ a voting layer that receives spikes from the spiking neurons of a fully connected (FC) layer to produce robust classification results~\cite{fang2021incorporating}.
The voting layer is implemented by a 1D average pooling layer with a kernel size of $K$ and a stride of $K$; in this study, we set $K=10$, as in the previous study~\cite{fang2021incorporating}.
The experimental results of these architectures by filling the TBD blocks with \texttt{SCB\_k3} and \texttt{SRB\_k3} are provided in \tablename~\ref{tab:discussion_on_macro_architecture} and \tablename~\ref{supp_tab:discussion_on_macro_architecture_SRB}, respectively.

\begin{table}[t]
\centering
\caption{Evaluation for different design choices on CIFAR10.}
\resizebox{\linewidth}{!}{
\addtolength{\tabcolsep}{-2pt}
\begin{tabular}{l|ccc|cc}
    \toprule
    Architecture & GAP & Normal & Down-sample & Acc.(\%) & Spikes \\
    \midrule
    SNN\_1 & \xmark & \texttt{SCB\_k3} & MaxPool & 86.93 & 154K \\
    \midrule
    SNN\_2 & \cmark & \texttt{SCB\_k3} & MaxPool & 85.05 & 168K \\
    \midrule
    SNN\_3 & \xmark & \texttt{SCB\_k3} & \texttt{SCB\_k3}  & 87.94 & 222K \\
    SNN\_4 & \xmark & \texttt{SCB\_k3} & AvgPool & 79.59 & 293K \\
    \bottomrule
\end{tabular}
\addtolength{\tabcolsep}{2pt}
}
\label{tab:discussion_on_macro_architecture}
\end{table}

\subsection{Use of Global Average Pooling Layer}

To observe the effect of the GAP layer~\cite{lin2013nin_gap}, we compare SNN\_1 and SNN\_2.
SNN\_2 includes a GAP layer with spiking neurons before the FC layer.
The GAP layer is commonly used to reduce the number of parameters of the FC layer in ANNs~\cite{szegedy2015inception,he2016resnet,sandler2018mobilenetv2} and SNNs~\cite{zheng2021going}.
However, our results indicate that the GAP layer has a negative effect on both the accuracy and energy efficiency.

\begin{figure}[t]
    \centering
    \includegraphics[width=\linewidth]{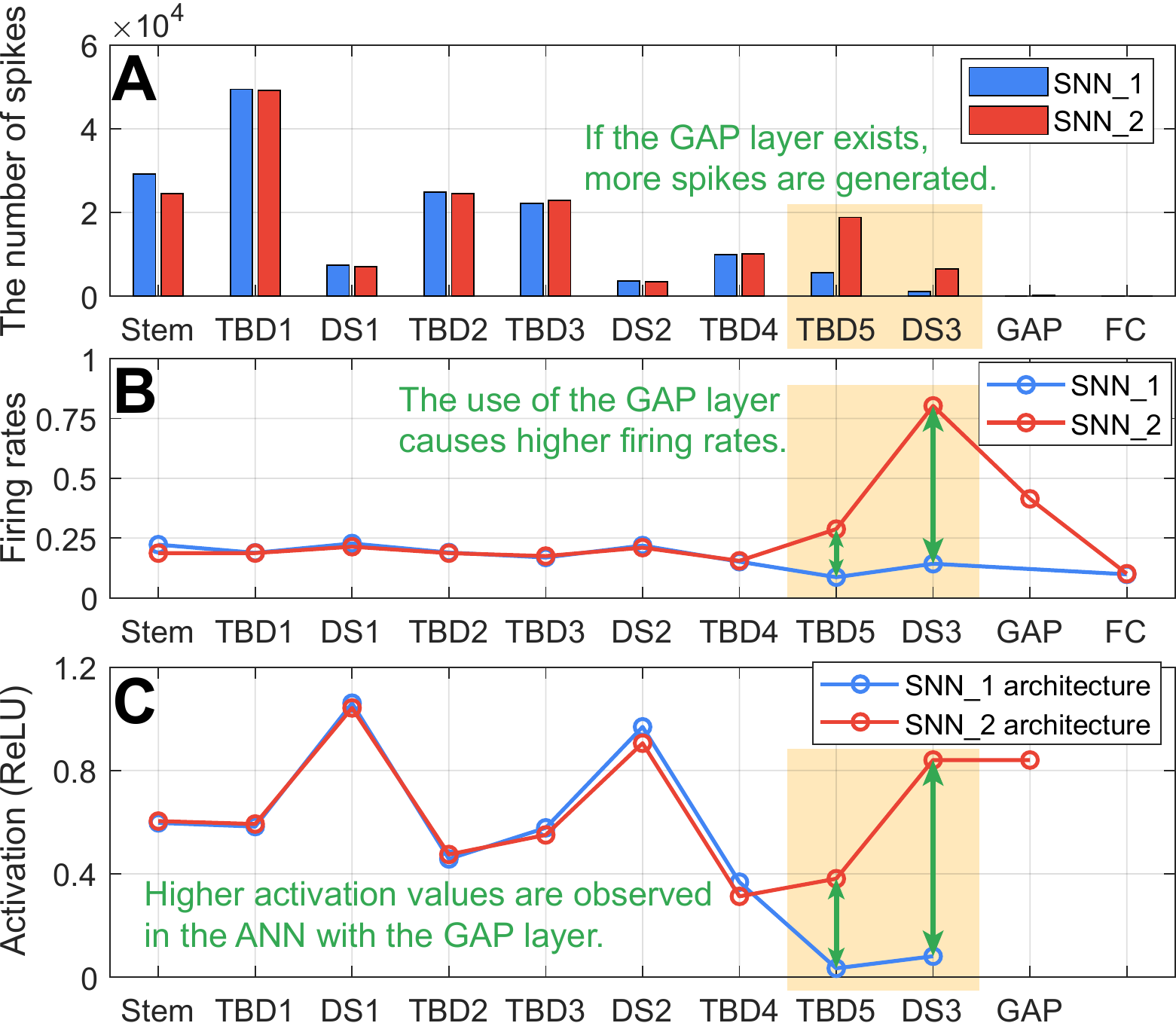}
    \vspace{-1em}
    \caption{Layerwise patterns of architectures without and with the GAP layer, denoted by SNN\_1 and SNN\_2, respectively. (A, B) The number of spikes and firing rates averaged over test data for 8 timesteps. (C) The average activation values in feature map of architectures, which replace spiking neurons with ReLU activation functions, averaged over test data.}
    \label{fig:snn1_vs_snn2}
\end{figure}

As presented in \tablename~\ref{tab:discussion_on_macro_architecture}, the accuracy of SNN\_2 is significantly lower than that of SNN\_1 and even more spikes are generated.
\figurename~\ref{fig:snn1_vs_snn2}A and B show the layerwise spike patterns in SNN\_1 and SNN\_2, where all the TBD blocks are \texttt{SCB\_k3}.
With the GAP layer, the number of spikes of the last TBD block (TBD5) and the last max pooling layer (DS3) significantly increases.
DS3 and the GAP layer have 4×4×4$C$ and 1×1×4$C$ spiking neurons, respectively, where $C$ is the initial channel of the architecture.
Hence, in SNN\_1 and SNN\_2, the input feature map sizes of the FC layer are 4×4×4$C$ and 1×1×4$C$, respectively. 
Consequently, the use of the GAP layer reduces the number of spiking neurons that transmit spike-based information to the FC layer.
To compensate for the information reduction caused by the GAP layer, the number of spikes and the firing rates of TBD5 and DS3 of SNN\_2 significantly increase.
Nevertheless, a non-negligible amount of information reduction occurs because of the reduced number of spiking neurons, leading to the observed accuracy drop.

When we replace spiking neurons with ReLU activation functions in these two architectures (\ie, ANN architectures) and train them, analogous phenomena are observed.
As shown in \figurename~\ref{fig:snn1_vs_snn2}C, for TBD5 and DS3, there are considerable differences in the average activation values between the two architectures.
In TBD5 and DS3, the increase in the average activation values for ANNs and average firing rates for SNNs may be caused by an architectural property related to the GAP layer.

Meanwhile, the firing rates in SNN\_2 and the activation values in the corresponding ANN behave differently.
When the membrane potential $V_\mathrm{mem}$ in the spiking neurons of the GAP layer does not exceed the threshold voltage $V_\mathrm{th}$, the remaining membrane potential cannot be transmitted to the following FC layer.
Consequently, as shown in \figurename~\ref{fig:snn1_vs_snn2}B, for SNN\_2, the firing rates of the GAP layer are lower than those of DS3.
This results in further information reduction, and thus the accuracy of SNN\_2 is lower than that of SNN\_1 (\tablename~\ref{tab:discussion_on_macro_architecture}).
In contrast, such information reduction does not occur in ANNs; the average activation values of the GAP layer do not differ from those of DS3, because by definition the GAP layer simply averages the activation values of DS3.
This is empirically supported by the fact that the accuracies of the two ANNs corresponding to SNN\_1 and SNN\_2 were approximately 90.5\%.
Therefore, unlike ANNs, when SNNs employ the GAP layer, the number of spikes increases and the information reduction occurs, suggesting that the GAP layer is an unsuitable design choice for performative and energy-efficient SNNs.

\subsection{Block Choice for Down-sampling}

In previous studies, different types of layers have been employed as down-sampling layers: max pooling layer~\cite{fang2021incorporating}, convolutional layer~\cite{zheng2021going}, and average pooling layer~\cite{wu2019direct,lee2020enabling}, which are used in SNN\_1, SNN\_3, and SNN\_4, respectively\footnote{The kernel sizes of these pooling layers are 2×2.}.
Thus, we investigate which down-sampling layer can lead to energy-efficient SNNs.
As shown in \tablename~\ref{tab:discussion_on_macro_architecture}, SNN\_3 with \texttt{SCB\_k3} achieves a 1\% higher accuracy than SNN\_1, because the use of trainable spiking blocks instead of max pooling layers increases the model capacity.
However, the number of spikes considerably increases by approximately 44\%.
SNN\_4 using average pooling layers experiences a significant drop in accuracy and also generates significantly increased spikes.
The information reduction discussed in the previous section is also likely to occur in average pooling layers.
Furthermore, we observed a large variance among the training results of SNN\_4 with different seeds; this implies that the use of average pooling layers leads to unstable training.
Hence, the use of trainable spiking blocks or average pooling layers is discouraged for the purpose of down-sampling.

\begin{figure}[t]
    \centering
    \includegraphics[width=\linewidth]{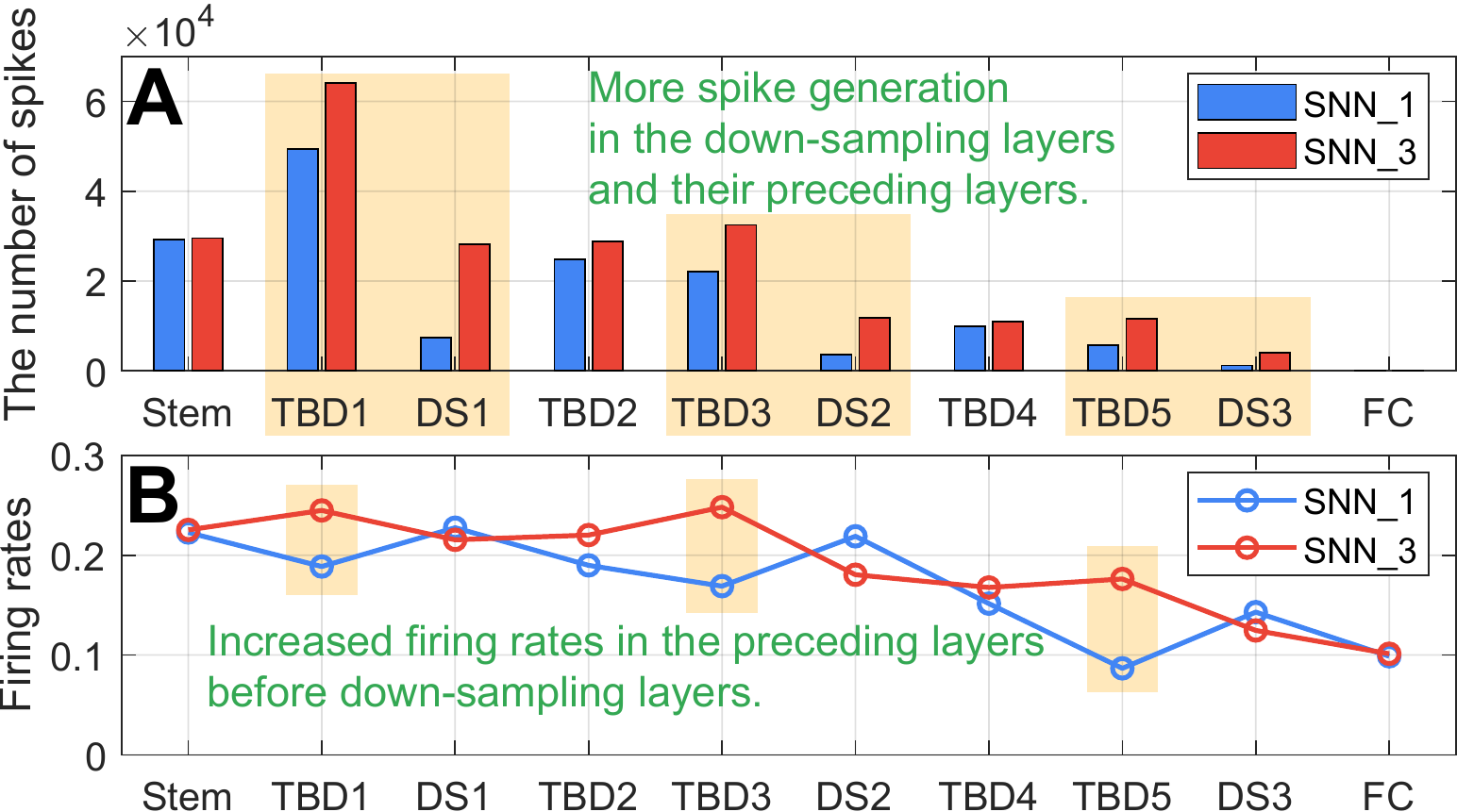}
    \vspace{-1em}
    \caption{Layerwise patterns of architectures that employ max pooling layers (SNN\_1) and \texttt{SCB\_k3} (SNN\_3) as down-sampling (DS) layers. (A) The number of spikes and (B) firing rates averaged over test data for 8 timesteps.}
    \label{fig:snn1_vs_snn3}
\end{figure}

We further compare the spike patterns of SNN\_1 and SNN\_3.
In \figurename~\ref{fig:snn1_vs_snn3}, the number of spikes in SNN\_3 increases not only in the down-sampling layers (\ie, DS1, DS2, and DS3) but also in their preceding layers (\ie, TBD1, TBD3, and TBD5).
The difference in the down-sampling layers is mainly caused by the difference between the size of feature map of the max pooling layer and the number of spiking neurons in \texttt{SCB\_k3}.
Thus, two convolutional layers with spiking neurons in \texttt{SCB\_k3} generate more spikes, even though the firing rates of max pooling and \texttt{SCB\_k3} are similar.
We now lay down the potential reason behind the increase in the firing rates in the preceding layers.
If a single input spike exists at least for the 2×2 kernel of the max pooling layer, this spike can be transmitted as the output spike.
Hence, the max pooling layers transmit information through the spikes more efficiently than \texttt{SCB\_k3}, and the preceding layers can generate fewer spikes without loss of information.
As a result, for energy-efficient SNNs, it is more desirable to use the max pooling layer for the down-sampling layers than the trainable spiking blocks.


\section{AutoSNN}\label{sec:AutoSNN}

In Section~\ref{sec:SNN_analysis}, we showed that excluding the GAP layer and using the max pooling layers effectively yields energy-efficient SNNs.
To further increase the energy efficiency, we leverage NAS, which automatically designs an optimal architecture.
We propose a spike-aware NAS framework, named AutoSNN, including both search space and search algorithm, which are described in Sections~\ref{subsec:search_space} and \ref{subsec:search_algorithm}, respectively.

\subsection{Energy-Efficient Search Space}\label{subsec:search_space}

It is important to define an expressive search space to effectively leverage NAS.
To this end, we define a search space on two levels: a macro-level backbone architecture and a micro-level candidate block set.
Based on the findings in Section~\ref{sec:SNN_analysis}, SNN\_1 is exploited as the macro-level backbone architecture, abbreviated as the macro architecture.
On the micro-level, we define a set of candidate spiking blocks derived from SCB and SRB with a kernel size $k$.
With $k \in \{3, 5\}$, the set of candidate blocks consists of five blocks: skip connection (\texttt{skip}), \texttt{SCB\_k3}, \texttt{SCB\_k5}, \texttt{SRB\_k3} and \texttt{SRB\_k5}.
These blocks are employed for each TBD block in the macro architecture.
With \texttt{skip}, search algorithms can choose to omit computation in certain TBD blocks.
We additionally considered a spiking block motivated by MobileNet~\cite{sandler2018mobilenetv2}, but this block was less suitable for designing energy-efficient SNNs than SCB and SRB; detailed discussion is provided in Section~\ref{supp:SIB_analysis}.

Herein, we evaluate the quality of the proposed search space in terms of the accuracy and number of spikes.
We compare four search spaces that consist of SNN architectures based on SNN\_$\{1, 2, 3, 4\}$ used in Section~\ref{sec:SNN_analysis}.
For each search space, we generate 100 architectures by randomly choosing blocks from the predefined candidate blocks to fill TBD blocks.
All the generated SNN architectures were trained on CIFAR10 for 120 epochs using a direct training method~\cite{fang2021incorporating}.

\begin{figure}[t]
    \centering
    \includegraphics[width=\linewidth]{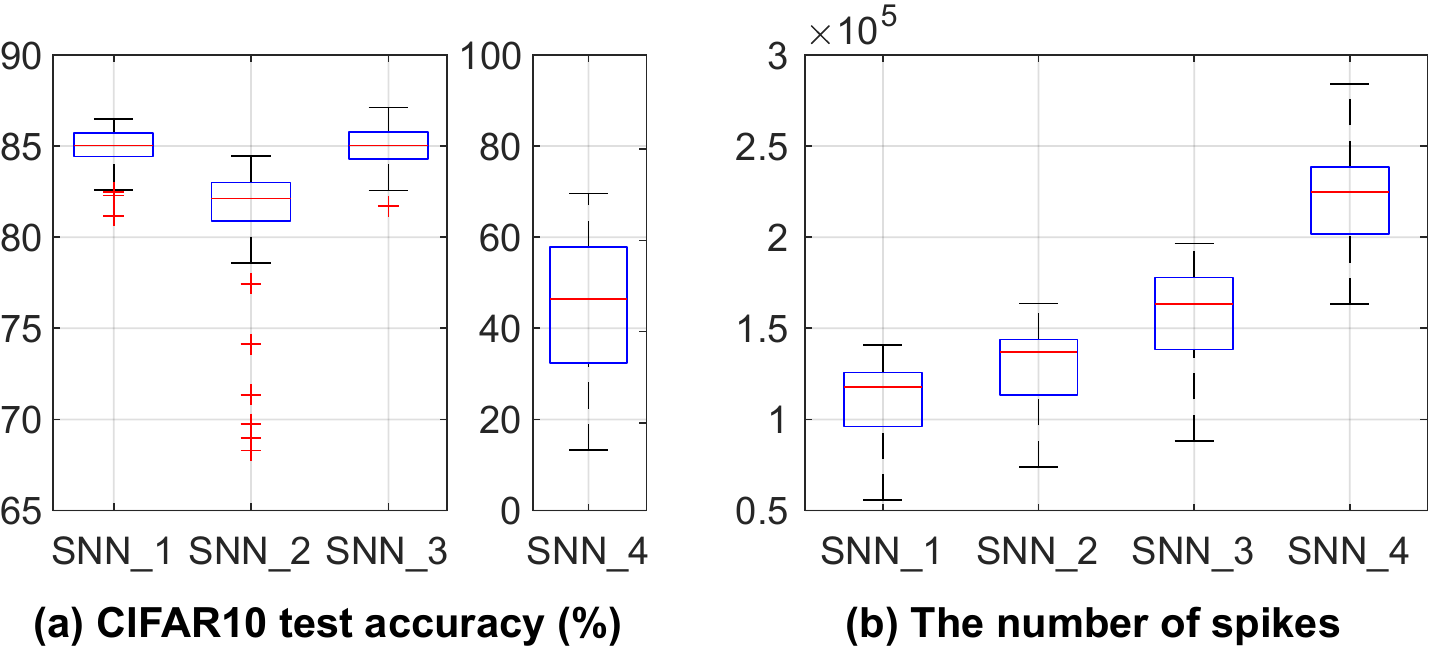}
    \vspace{-1em}
    \caption{Search space quality comparison between SNN\_1-based (proposed) and SNN\_$\{2,3,4\}$-based search spaces.}
    \label{fig:search_space_quality}
\end{figure}

In \figurename~\ref{fig:search_space_quality}, the search space quality comparison empirically validates that the proposed search space based on SNN\_1 is of higher quality than its variations in terms of the accuracy and number of spikes. 
This result is consistent with the analysis presented in Section~\ref{sec:SNN_analysis}.
The proposed search space includes architectures with higher accuracy and fewer spikes on average than those in the SNN\_2-based search space.
A similar but more distinctive pattern is observed in the SNN\_4-based search space.
Once again, this indicates that the use of the GAP layer and average pooling layers is inappropriate for finding performative and energy-efficient SNNs.
In the SNN\_3-based search space, the accuracy is comparable but the average number of spikes increases by approximately x1.4 over the proposed search space.
Hence, the excessive use of spiking blocks with trainable parameters (\ie, SCB and SRB) also needs to be avoided.

\subsection{Spike-aware Search Algorithm}\label{subsec:search_algorithm}

In general, according to Eq.~\ref{eq:LIF_dynamics_discrete} and Section~\ref{supp:direct_training}, SNNs with timesteps take longer to train than ANNs requiring feed-forward and backpropagation once.
Thus, reducing the search cost induced by training and evaluating candidate architectures becomes even more critical in NAS for SNNs than for ANNs.
To address this challenge, we adopt a one-shot weight-sharing approach based on an evolutionary algorithm.
AutoSNN consists of two consecutive procedures, which are illustrated in \figurename~\ref{fig:overview_AutoSNN}: 1) training a super-network that encodes the proposed search space and 2) evaluating candidate architectures under a search budget to search for an optimal architecture.
In AutoSNN, the target dataset is divided into training data $D_\mathrm{train}$ and validation data $D_\mathrm{val}$, which are used for the first and second procedures, respectively.

In the first procedure, AutoSNN trains the architectures sampled from a super-network whose subnetworks correspond to all the candidate architectures in the proposed search space.
To evenly train spiking blocks in the super-network, we adopt the single-path uniform sampling~\cite{bender2018understanding,li2020random,guo2020spos,zhang2021rlnas,yan2021lighttrack} because of its effectiveness and simplicity.
Based on the supervised training method of SNN~\cite{fang2021incorporating}, each sampled architecture is trained on a single mini-batch from $D_\mathrm{train}$.

Given a trained super-network, to enable a spike-aware search, AutoSNN penalizes SNNs with more spikes.
To evaluate SNN architectures, we define the architecture fitness $F(A)$ based on exponential discounting as follows:
\begin{equation}
    F(A) = Accuracy \times (N / N_\mathrm{avg})^{\lambda},
\end{equation}
where $N$ is the number of spikes generated by architecture $A$, $N_\mathrm{avg}$ is the average number of spikes across architectures sampled during training the super-network, and $\lambda$ is the coefficient that controls the influence of the spike-aware term.
We set $\lambda < 0$ to explicitly search for architectures with lower $N$.
As $|\lambda|$ increases, the SNNs discovered by AutoSNN are expected to generate fewer spikes.
Note that other discounting functions are viable.
Using a fitness function with linear discounting ($F'(A) = Acc. - \lambda' (N/N_\mathrm{avg})$) led to similar results to those using our exponential discounting function.

Using the obtained fitness value of each candidate architecture, AutoSNN based on an evolutionary algorithm explores the proposed search space and finds the architecture with the highest fitness value.
We briefly describe the evolutionary search algorithm; a detailed explanation is provided in Section~\ref{supp:search_algorithm} along with Algorithm~\ref{alg:search_algorithm}.
AutoSNN maintains two population pools throughout the search process: the top-$k$ population pool P$_\textrm{top}$ and the temporary evaluation population pool P$_\textrm{eval}$.
First, P$_\textrm{eval}$ is prepared by generating architectures using evolutionary techniques: mutation and crossover.
For these techniques, the parent architectures are sampled from P$_\textrm{top}$.
The architectures in P$_\textrm{eval}$ are evaluated using the proposed spike-aware fitness and $D_\mathrm{val}$.
If the fitness values of the evaluated architectures are higher than those of the architectures in P$_\textrm{top}$, P$_\textrm{top}$ is updated.
These processes are repeated until $B$ architectures are evaluated, where $B$ is a search budget; in this study, we set $B$ as 200.
Finally, AutoSNN obtains architecture $A^\ast$ with the highest fitness value from \textrm{P$_\textrm{top}$}.

Because the two procedures are decoupled, AutoSNN can discover a promising SNN architecture for every different neuromorphic chip by simply changing $\lambda$.
Similar to a previous study~\cite{cai2020ofanet}, AutoSNN can reuse a trained super-network and execute the second procedure alone.
Unlike the differentiable NAS approach, where the entire search process is executed to find a single architecture, the additional search cost is negligible, as demonstrated in Section~\ref{sec:results}. 
Therefore, our search algorithm based on two separate procedures is a practical and effective method for obtaining energy-efficient SNNs.

\section{Experiments and Discussion}\label{sec:results}

\subsection{Experimental Settings}\label{subsec:setting}

We evaluated the SNNs searched by AutoSNN on two types of datasets: static datasets (CIFAR10, CIFAR100~\cite{krizhevsky2009cifar10}, SVHN~\cite{netzer2011svhn}, and Tiny-ImageNet-200\footnote{https://www.kaggle.com/akash2sharma/tiny-imagenet}) and neuromorphic datasets (CIFAR10-DVS~\cite{li2017cifar10dvs} and DVS128-Gesture~\cite{amir2017dvs128gesture}).
Details regarding these datasets are provided in Section~\ref{supp:dataset}.
The dataset is divided into 8:2 for $D_\mathrm{train}$ and $D_\mathrm{val}$.
We use the Adam optimizer~\cite{kingma2015adam} with a learning rate of 0.001 and cutout data augmentation~\cite{devries2017cutout} to train the super-network and the searched SNNs for 600 epochs on a single NVIDIA 2080ti GPU.
For all architectures, we use PLIF neurons~\cite{fang2021incorporating} with $V_\mathrm{th}=0$, $V_\mathrm{reset}=0$, 8 timesteps, and an initial $\tau$ of 2.

\subsection{Searching with Different $\lambda$ of Spike-aware Fitness}

By varying $\lambda \in \{0, -0.08, -0.16, -0.24\}$, we execute AutoSNN and report the results in \tablename~\ref{tab:lambda_change}; the searched architectures are visualized in \figurename~\ref{supp_fig:searched_arch_lambda}.
Note that on a single 2080ti GPU, the search cost is approximately 7 GPU hours: 6 h 48 min for training a super-network and 8 min for executing the evolutionary search.
For each $\lambda$, it incurs a negligible additional cost (8 min).

\begin{table}[t]
\centering
\caption{Searching results with different $\lambda$ on CIFAR10.}
\begin{tabular}{l|cccc}
    \toprule
    $\lambda$ in fitness & 0 & $-0.08$ & $-0.16$ & $-0.24$ \\
    \midrule
    Acc (\%) $\uparrow$ & 88.69 & 88.67 & 88.46 & 86.58 \\
    Spikes $\downarrow$ & 127K & 108K & 106K & 54K \\
    \bottomrule
\end{tabular}
\label{tab:lambda_change}
\end{table}

In \figurename~\ref{supp_fig:searched_arch_lambda}, distinctive differences among the SNN architectures searched with different $\lambda$ values are observed; detailed discussion is provided in Section~\ref{supp:different_lambda}.
Increasing $|\lambda|$ leads to architectures with more TBD blocks filled with \texttt{Skip}, thereby decreasing the number of spikes.
This confirms that $\lambda$ functions according to our intent to adjust the trade-off between the accuracy and number of spikes in the searched SNN.
Compared to the architecture with $\lambda=0$, the one searched with $\lambda=-0.08$ reduces approximately 20K spikes while achieving a similar accuracy.
Thus, in other experiments, we use $\lambda=-0.08$ by default.

\subsection{Comparison with Existing SNNs}

\begin{table}[t]
\centering
\caption{Evaluation of SNN architectures with an initial channel of $C$ on CIFAR10. $^\dagger$ denotes the initial channel used in previous studies. $^\ddagger$ denotes reported values in original papers (\ie, not reproduced in our training settings).}
\renewcommand{\arraystretch}{0.9}
\addtolength{\tabcolsep}{-2pt}
\begin{tabular}{lr|rrr}
    \toprule
    \multirow{2}{*}{SNN Architecture} & \multirow{2}{*}{$C$} & Acc & Spikes & Params \\
     & & (\%) $\uparrow$ & $\downarrow$ & (M) \\
    \midrule
    CIFARNet-Wu & 16    & 84.36  & 361K & 0.71 \\
    \cite{wu2019direct} & 32    & 86.62 & 655K & 2.83 \\
    & 64    & 87.80  & 1298K & 11.28 \\
    & $^\dagger$128   & $^\ddagger$90.53 &  - & 45.05\\
    \midrule
    CIFARNet-Fang & 16 & 80.82 & 104K & 0.16 \\
    \cite{fang2021incorporating} & 32	& 86.05	& 160K & 0.60\\
     & 64	& 90.83	& 260K & 2.34\\
    & 128	& 92.33	& 290K & 9.23 \\
    & $^\dagger$256   & 93.15 & 507K & 36.72 \\
    \midrule
    ResNet11-Lee & 16	& 84.43	& 140K & 1.17 \\
    \cite{lee2020enabling} & 32	& 87.95	& 301K & 4.60 \\
    & $^\dagger$64    & $^\ddagger$90.24 & $^\ddagger$1530K & 18.30 \\
    \midrule
    ResNet19-Zheng & 16	& 83.95	& 341K & 0.23 \\
    \cite{zheng2021going} & 32	& 89.51	& 541K & 0.93 \\
    & 64	& 90.95	& 853K & 3.68 \\
    & $^\dagger$128   & 93.07 & 1246K & 14.69 \\
    \midrule
    \textbf{AutoSNN}  & 16 & 88.67 & 108K & 0.42\\
    (proposed)        & 32 & 91.32 & 176K & 1.46\\
             & 64 & 92.54 & 261K & 5.44\\
             & 128 & 93.15 & 310K & 20.92\\
    \bottomrule
\end{tabular}
\addtolength{\tabcolsep}{2pt}
\label{tab:main_search_results}
\end{table}

When AutoSNN is executed, an initial channel of 16 is used for the macro architecture, but the hand-crafted SNNs used in previous studies had a noticeably larger number of initial channels denoted by $^\dagger$ in \tablename~\ref{tab:main_search_results}.
Hence, for a fair comparison, we increase and decrease the initial channel by a factor of two, for AutoSNN and the other SNNs, respectively.
We train all SNNs under the same training settings.
The evaluation results for CIFAR10 are provided in \tablename~\ref{tab:main_search_results}; refer to \figurename~\ref{fig:spikes_vs_acc} for the visual summary.

\begin{figure}[t]
    \centering
    \includegraphics[width=\linewidth]{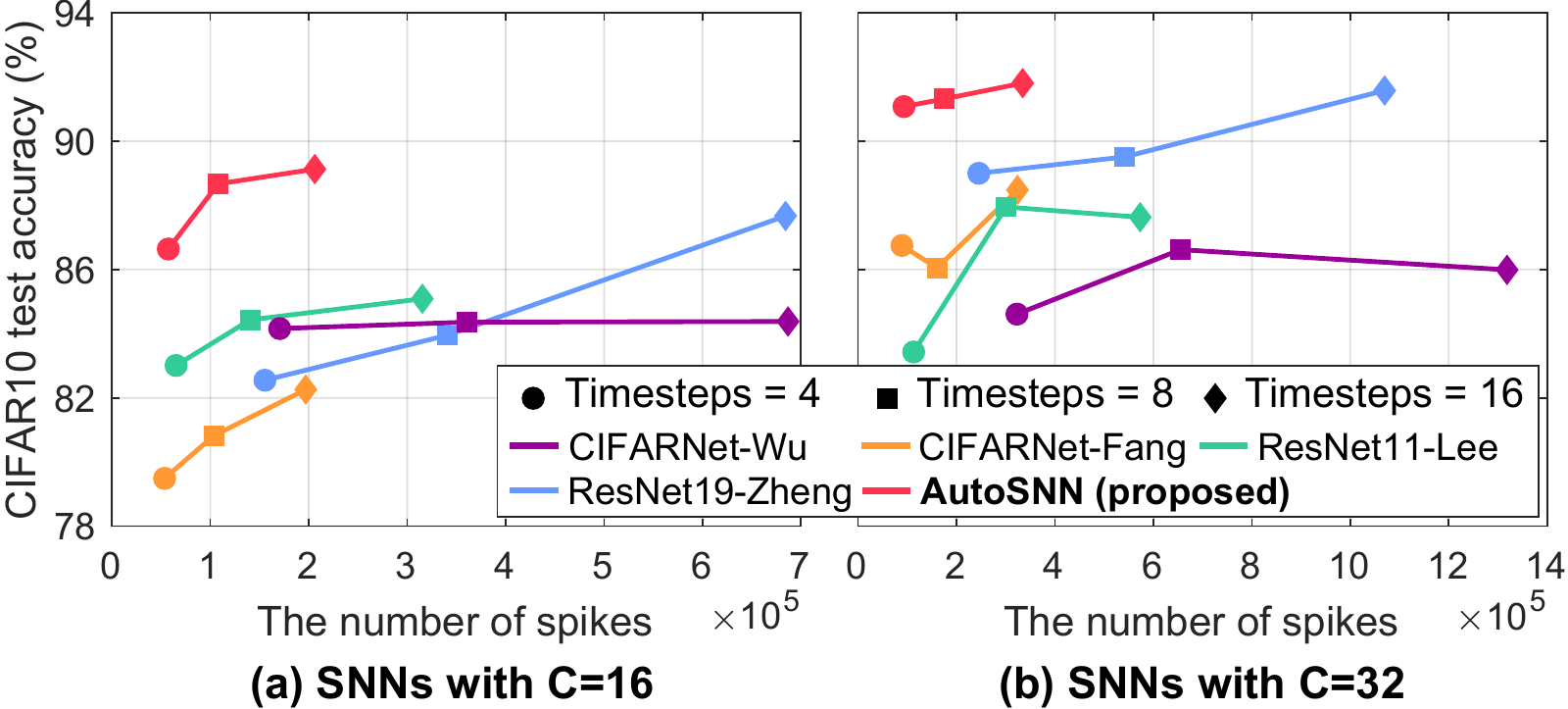}
    \vspace{-1em}
    \caption{The number of spikes vs. CIFAR10 accuracy of various SNN architectures trained with different timesteps $\in \{4, 8, 16\}$.}
    \label{fig:spikes_vs_acc_different_timesteps}
\end{figure}

The experimental results confirm that AutoSNN successfully discovers an energy-efficient SNN, which lies on the Pareto front in the region of accuracy and the number of spikes.
In \figurename~\ref{fig:spikes_vs_acc_different_timesteps}, the superiority of AutoSNN is also observed even when training SNNs with different timesteps, 4 and 16.
This demonstrates that the hand-crafted SNNs have undesirable architectural components and the NAS-based approach can be essential to improve energy efficiency.

We further evaluate our searched SNN architecture by transferring it to various datasets.
For the three static datasets, an initial channel is set to 64 to account for the increased complexity of these datasets.
For datasets with a larger resolution than CIFAR10 (\ie, Tiny-ImageNet-200, CIFAR10-DVS, and DVS128-Gesture), we use macro architectures with deeper stem layers, which are visualized in \figurename~\ref{supp_fig:macro_for_dvs_data}.
\tablename~\ref{tab:different_datasets} clearly reveals that AutoSNN achieves higher accuracy and generates fewer spikes than the other hand-crafted architectures across all datasets including both static and neuromorphic datasets.

\begin{table}[t]
\centering
\caption{Evaluation of SNN architectures with the initial channel of $C$ on various datasets. $^\dagger$ denotes architectures with additional stem layers. $^\ddagger$ denotes reported values in original papers.}
\resizebox{\columnwidth}{!}{
\addtolength{\tabcolsep}{-1.5pt}
\begin{tabular}{llr|rr}
    \toprule
    Data & SNN Architecture & $C$ & Acc (\%) $\uparrow$ & Spikes $\downarrow$ \\
    \midrule
    CIFAR100 & Fang et al. 2021 & 256 & 66.83 & 716K \\
    & \textbf{AutoSNN} & 64 & 69.16 & 326K \\
    \midrule
    SVHN & Fang et al. 2021 & 256 & 91.38 & 462K \\
    & \textbf{AutoSNN} & 64 & 91.74 & 215K \\
    \midrule
    Tiny-Image & Fang et al. 2021 & 256 & 45.43 & 1724K \\
    Net-200 & \textbf{AutoSNN}$^\dagger$ & 64 & 46.79 & 680K \\
    \midrule
    CIFAR10 & Wu et al. 2019b & 128 & $^\ddagger$60.50 & - \\ 
    -DVS & Fang et al. 2021 & 128 & 69.10 & 4521K \\
     & Zheng et al. 2021 & 64 & 66.10 & 1550K \\ 
    & \textbf{AutoSNN}$^\dagger$ & 16 & 72.50 & 1269K \\
    \midrule
    DVS128 & He et al. 2020 & 64 & $^\ddagger$93.40 & - \\
    -Gesture & Kaiser et al. 2020 & 64 & $^\ddagger$95.54 & - \\
    & Fang et al. 2021 & 128 & 95.49 & 1459K \\
    & Zheng et al. 2021 & 64 & 96.53 & 1667K \\ 
    & \textbf{AutoSNN}$^\dagger$ & 16 & 96.53 & 423K \\
    \bottomrule
\end{tabular}
\addtolength{\tabcolsep}{1.5pt}
}
\label{tab:different_datasets}
\end{table}

\subsection{Validity of the Search Algorithm in AutoSNN}

\begin{table}[t]
\centering
\caption{Ablation study results of AutoSNN on CIFAR10. WS is a shorthand for weight-sharing.}
\addtolength{\tabcolsep}{-1pt}
\renewcommand{\arraystretch}{0.9}
\begin{tabular}{lrr}
    \toprule
    Search & Acc (\%) $\uparrow$ & Spikes $\downarrow$ \\
    \midrule
    Random sampling  & 86.97$\pm$1.06 & 123K$\pm$29K\\
    \midrule
    \multicolumn{3}{l}{WS + random search}  \\
    ~~$\lambda = 0$ & 88.40 & 132K \\
    ~~$\lambda = -0.08$ (spike-aware) & 88.10 & 133K \\
    \midrule
    \multicolumn{3}{l}{WS + evolutionary search (AutoSNN)}  \\
    ~~$\lambda = 0$ & 88.69 & 127K \\
    ~~$\lambda = -0.08$ (spike-aware) & 88.67 & 108K \\
    \bottomrule
\end{tabular}
\addtolength{\tabcolsep}{1pt}
\label{tab:search_effectiveness_ablation}
\end{table}

\textbf{Ablation Study for Two Procedures.}
Through an ablation study in \tablename~\ref{tab:search_effectiveness_ablation}, we inspect two procedures in the search algorithm of AutoSNN.
First, we randomly sample and train 10 architectures (Random sampling); the cost is approximately 100 GPU hours.
Second, using a trained super-network, we select the architecture with the highest spike-aware fitness among the 200 randomly-sampled architectures (WS + random search), which is the same search budget as AutoSNN.
As shown in \tablename~\ref{tab:search_effectiveness_ablation}, SNNs searched by WS + random search yield higher accuracy than the average accuracy of 10 randomly-sampled architectures.
This indicates that the weight-sharing strategy with a direct training method of SNNs is valid in the SNN domain.
Applying evolutionary search further improves the search result, solidifying the effectiveness of our evolutionary search with spike-aware fitness.

\begin{table}[t]
\centering
\caption{Evaluation for AutoSNN on enlarged search spaces with eight TBD blocks ($C=16$).}
\resizebox{\columnwidth}{!}{
\renewcommand{\arraystretch}{0.9}
\addtolength{\tabcolsep}{-2pt}
\begin{tabular}{lrr}
    \toprule
    SNN Architecture & Acc (\%) $\uparrow$ & Spikes $\downarrow$\\
    \midrule
    \multicolumn{3}{l}{Using TBD blocks instead of max pooling layers} \\
    \texttt{SCB\_k3} in all TBD blocks & 87.94 & 222K \\
    \texttt{SRB\_k3} in all TBD blocks & 89.18 & 221K \\
    AutoSNN ($\lambda=-0.04$) & 89.05 & 170K \\
    AutoSNN ($\lambda=-0.08$) & 87.92 & 65K \\
    \midrule
    \multicolumn{3}{l}{Adding a TBD block before each max pooling layer} \\
    \texttt{SCB\_k3} in all TBD blocks & 87.04 & 230K \\
    \texttt{SRB\_k3} in all TBD blocks & 88.69 & 228K \\
    AutoSNN ($\lambda=-0.08$) & 88.60 & 143K \\
    AutoSNN ($\lambda=-0.16$) & 87.29 & 60K \\
    \bottomrule
\end{tabular}
\addtolength{\tabcolsep}{-2pt}
}
\label{tab:larger_search_space}
\end{table}

\textbf{Searching on Enlarged Search Spaces.}
AutoSNN is effective in searching for desirable architectures in enlarged search spaces.
The proposed SNN search space consists of 3,125 architectures ($5^5$; five candidate blocks and five TBD blocks).
We construct two search spaces, where the macro architectures have eight TBD blocks as described in \tablename~\ref{tab:larger_search_space}; both of them include 390,625 architectures ($5^8$).
One is the SNN\_3-based search space, and the other consists of architectures where a TBD block is added before each max pooling layer of SNN\_1.
\tablename~\ref{tab:larger_search_space} shows the validity of the search algorithm of AutoSNN.
In both search spaces, compared with architectures consisting of \texttt{SCB\_k3} and \texttt{SRB\_k3}, AutoSNN discovers architectures which generate significantly fewer spikes.

\subsection{Architecture Search without Spiking Neurons}\label{subsec:ANN_search}

\begin{table}[t]
\centering
\caption{Searching results on CIFAR10 from ANN and SNN search spaces, where $\lambda$ in fitness is set as 0 for a fair comparison.}
\begin{tabular}{lrr}
    \toprule
    Search space  & Acc (\%) $\uparrow$ & Spikes $\downarrow$ \\
    \midrule
    w/o spiking neurons (ANN) & 88.02 & 134K \\
    w/ spiking neurons (proposed) & 88.69 & 127K \\
    \bottomrule
\end{tabular}
\label{tab:ann_search}
\end{table}

To validate the importance of considering the properties of SNNs during the search process, we execute our evolutionary search algorithm on an ANN search space, where spiking neurons in architectures are removed; instead, ReLU activation functions are used.
Because spikes cannot be observed in the ANNs, the accuracy is solely used to evaluate the architectures.
After finding an architecture from the ANN search space, spiking neurons are added to this architecture, which is then trained according to the settings in Section~\ref{subsec:setting}.
As presented in \tablename~\ref{tab:ann_search}, the SNN architecture searched by AutoSNN with $\lambda = 0$ achieves higher accuracy and generates fewer spikes than the architecture searched from the ANN search space.
We conjecture that this discrepancy because training the super-network without spiking neurons cannot reflect the properties of SNNs such as spike-based neural dynamics to represent information.

\section{Conclusion}

For energy-efficient artificial intelligence, it is essential to design SNNs that have minimal spike generation and yield competitive performance. 
In this study, we proposed a spike-aware fitness and AutoSNN, a spike-aware NAS framework to effectively search for such SNNs from the energy-efficient search space that we defined.
To define the search space, we analyzed the effects of the architecture components on the accuracy and number of spikes.
Based on our findings, we suggested excluding the GAP layer and employing the max pooling layers as down-sampling layers in SNNs.
From the search space that consists of SNN architectures satisfying these design choices, AutoSNN successfully discovered the SNN architecture that is most performative and energy-efficient compared with various architectures used in previous studies.
Our results highlighted the importance of architectural configurations in the SNN domain.
We anticipate that this study will inspire further research into automatic design of energy-efficient SNN architectures.

\section*{Acknowledgment}

This work was supported by the National Research Foundation of Korea (NRF) grant funded by the Korea government (MSIT) [No. 2022R1A3B1077720, No. 2021R1C1C2010454], Institute of Information \& communications Technology Planning \& Evaluation (IITP) grant funded by the Korea government (MSIT) [NO. 2021-0-01343, Artificial Intelligence Graduate School Program (Seoul National University)], and the Brain Korea 21 Plus Project in 2022.


\bibliography{references}
\bibliographystyle{icml2022}

\newpage
\appendix


\section{Experimental Environment}

\subsection{Experimental Settings}

We implemented AutoSNN and all the experiments using SpikingJelly\footnote{https://github.com/fangwei123456/spikingjelly}, and have included the codes in the supplementary materials (code.zip).
The code will be made public on github after the review process.

For all the experiments, we used a single GeForce RTX 2080 Ti GPU and a supervised training method that utilizes PLIF neurons with an initial $\tau$ of 2 (\ie, $\alpha=-ln(1)$), $V_\mathrm{reset}=0$, and $V_\mathrm{th}=1$~\cite{fang2021incorporating}.
We evaluated SNN architectures on two types of datasets: static datasets (CIFAR-10~\cite{krizhevsky2009cifar10}, CIFAR-100~\cite{krizhevsky2009cifar10}, SVHN~\cite{netzer2011svhn}, and Tiny-ImageNet-200\footnote{https://www.kaggle.com/akash2sharma/tiny-imagenet}) and neuronmorphic datasets (CIFAR10-DVS~\cite{li2017cifar10dvs} and DVS128-Gesture~\cite{amir2017dvs128gesture}).
Details regarding these datasets are provided in Section~\ref{supp:dataset}.
We set timesteps as 8 and 20 for the static datasets and the neuromorphic datasets, respectively.
When executing AutoSNN to search for SNNs on the proposed search space, CIFAR10 was used.
The training data of CIFAR10 were divided into 8:2 for $D_\mathrm{train}$ and $D_\mathrm{val}$, which were used to train the super-network and evaluate candidate architectures during the spike-aware evolutionary search, respectively.

\subsubsection{Training an SNN-based Super-network}
To train the super-network, we employed the Adam optimizer~\cite{kingma2015adam} with a fixed learning rate of 0.001 and a momentum of (0.9, 0.999).
The super-network was trained for 600 epochs with a batch size of 96.
During the training, we applied three conventional data preprocessing techniques: the channel normalization, the central padding of images to 40×40 and then random cropping back to 32×32, and random horizontal flipping.

\subsubsection{Spike-aware Evolutionary Search}
Once the super-network is trained, AutoSNN can evaluate candidate architectures that are sampled from the search space or generated by mutation and crossover.
The architectures inherit the weights of the trained super-network.
We set the search algorithm parameters as follows: a maximum round $\mathcal{T}$ of 10, a mutation probability $\rho$ of 0.2, the number of architectures generated by mutation $p_m$ of 10, the number of architectures generated by crossover $p_c$ of 10, the size of the top-$k$ architecture pool $k$ of 10, and the size of the evaluation pool $p$ of 20.

\subsubsection{Training SNN Architectures}
\begin{figure}
    \centering
    \includegraphics[width=\linewidth]{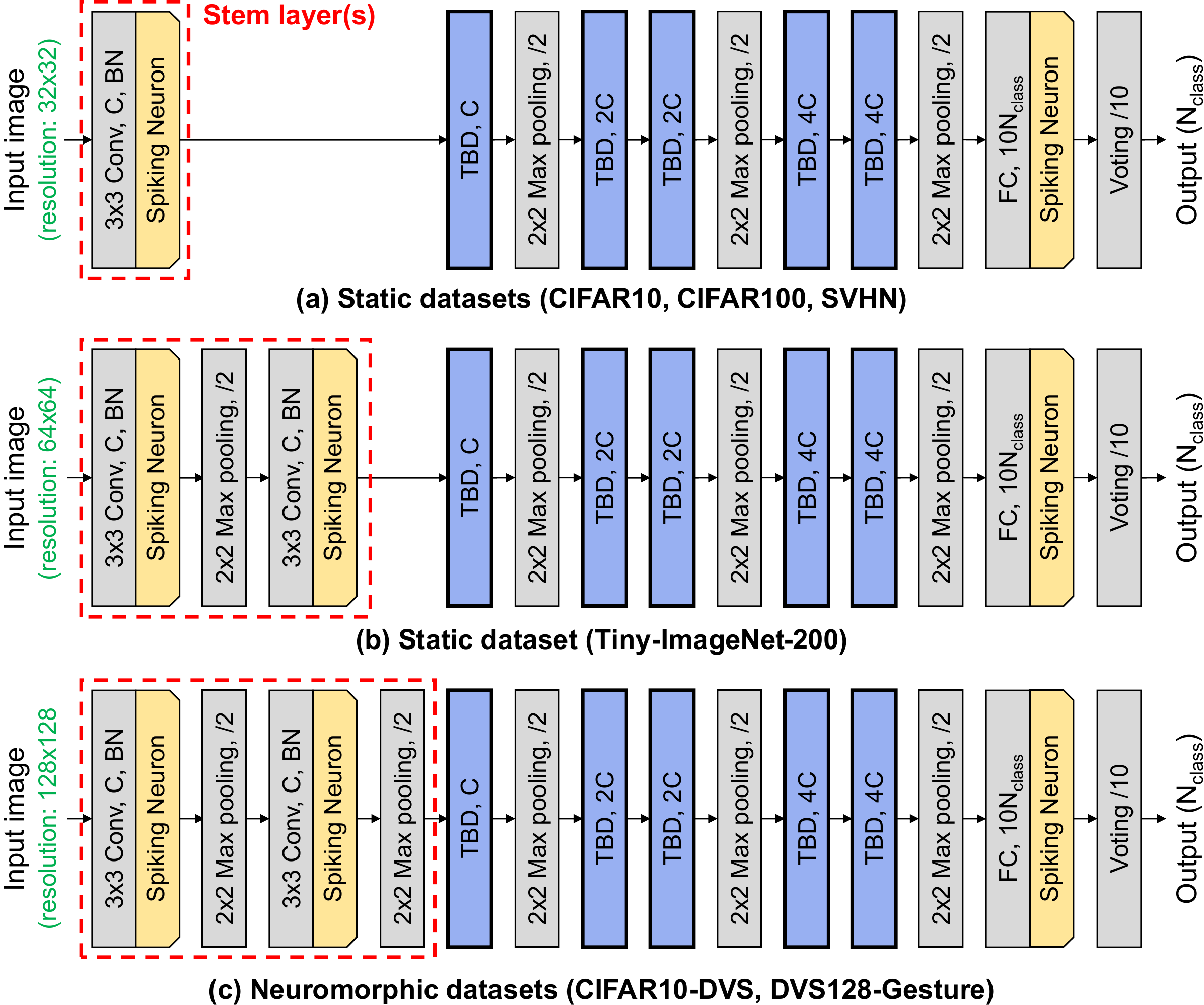}
    \vspace{-1em}
    \caption{The macro architectures used for various datasets: (a) CIFAR-10, CIFAR-100, and SVHN with a resolution of 32×32, (b) Tiny-ImageNet-200 with a resolution of 64×64, and (c) CIFAR10-DVS and DVS128-Gesture with a resolution of 128×128. To reduce the resolution of image into 32×32, architectures for Tiny-ImageNet-200 and neuromorphic datasets include additional layers in stem layers.}
    \label{supp_fig:macro_for_dvs_data}
\end{figure}

All the SNNs including architectures searched by AutoSNN and conventional hand-crafted architectures were trained from scratch for 600 epochs with a batch size of 96.
Some SNNs, where the batch size of 96 was not executable due to the memory size, were trained with a smaller batch size.
We also used the Adam optimizer~\cite{kingma2015adam} with a fixed learning rate of 0.001 and a momentum of (0.9, 0.999).
For the static datasets, we additionally applied cutout data augmentation with length 16~\cite{devries2017cutout}, along with the three conventional data augmentation techniques applied in the super-network training.
When training Tiny-ImageNet-200, a max pooling and a 3×3 convolution layer are sequentially added into the stem layer of our macro architecture backbone to reduce the resolution from 64×64 to 32×32, as shown in \figurename~\ref{supp_fig:macro_for_dvs_data}(b).
For the neuromorphic datasets with a resolution of 128×128, we constructed deeper stem layer in order to reduce the resolution from 128×128 to 32×32 such that the subsequent layers searched by AutoSNN can process the data, as shown in \figurename~\ref{supp_fig:macro_for_dvs_data}(c).

\subsection{Dataset Description}\label{supp:dataset}

\subsubsection{Static Datasets}
CIFAR-10, CIFAR-100, and SVHN include images with a resolution of 32×32 and 3 channels (RGB channels).
An image corresponds to one static frame with pixel values, and thus we refer to these datasets as the static datasets.
CIFAR-10 and CIFAR-100 are composed of 50,000 training data and 10,000 test data, while SVHN has approximately 73K images for training and 26K images for testing.
Tiny-ImageNet-200 includes images with a resolution of 64×64 and 3 channels, where the images are sampled from ImageNet~\cite{russakovsky2015imagenet} and downsized from 224x224 to 64×64.
100K images and 2.5K images are used for training and testing, respectively.
These datasets are used in the classification task: 10 classes for CIFAR-10 and SVHN, 100 classes for CIFAR-100, and 200 classes for Tiny-ImageNet-200.

\subsubsection{Neuromorphic Datasets}
We evaluate SNN architectures on the neuromorphic datasets, which include data with a format of event stream, referred to as a spike train in our context.
The datasets are collected by using a dynamic vision sensor (DVS), which outputs 128×128 images with 2 channels. 
For CIFAR10-DVS~\cite{li2017cifar10dvs}, 10,000 images from CIFAR-10 are converted into the spike trains, and there are 1,000 images per class.
We divide the dataset into two parts: 9,000 images for the training data and 1,000 images for the test data.
DVS128-Gesture~\cite{amir2017dvs128gesture} includes 1,342 training data and 288 test data with 11 classes of hand gestures.

\section{Supplemental Material of Section~\ref{sec:SNN_analysis}}\label{supp:snn_analysis}

\subsection{Architecture Preparation for Analysis}

\begin{figure}[t]
    \centering
    \includegraphics[width=\linewidth]{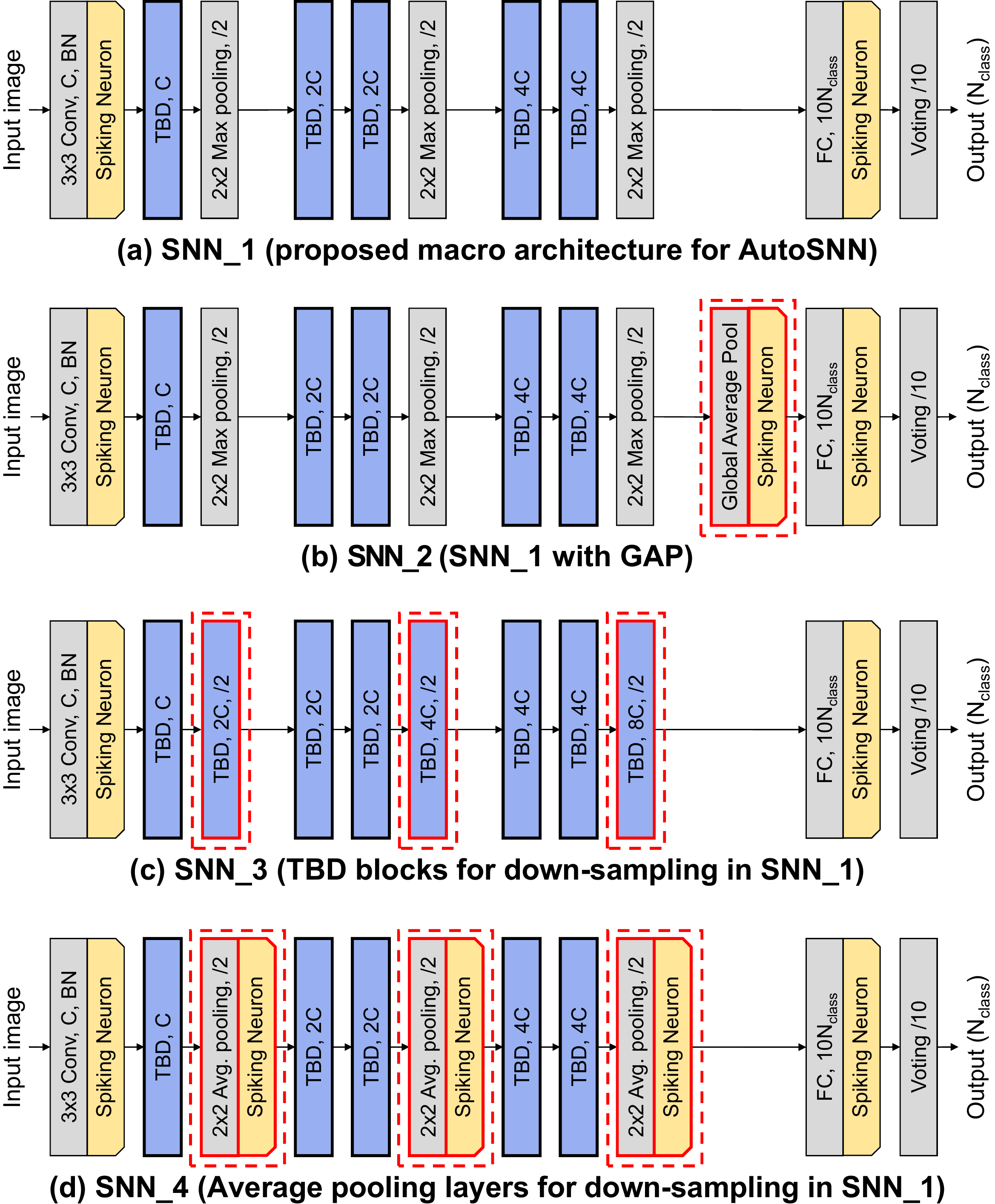}
    \vspace{-1em}
    \caption{Architectures for analyzing the architectural effects: (a) SNN\_1 and (b-d) its variants, SNN\_$\{2, 3, 4\}$. The red dotted boxes indicate the change from SNN\_1.}
    \label{supp_fig:macro_vs_variants}
\end{figure}

In Section~\ref{sec:SNN_analysis}, we analyzed the architectural effects on accuracy and the number of spikes of SNNs.
For the analysis, motivated by architectures used in the previous studies~\cite{lee2020enabling,fang2021incorporating,zheng2021going}, we prepared architecture variations, \ie, SNN\_$\{1, 2, 3, 4\}$, which are depicted in \figurename~\ref{supp_fig:macro_vs_variants}.
SNN\_1 is a base architecture, SNN\_2 includes the GAP layer, and SNN\_3 and SNN\_4 use trainable spiking blocks and average pooling layers as down-sampling layers, respectively.

\subsection{Filling TBD Blocks with \texttt{SRB\_k3}}

\tablename~\ref{supp_tab:discussion_on_macro_architecture_SRB} presents additional results of the architectures, where all TBD blocks are solely filled with \texttt{SRB\_k3}.
As discussed in Section~\ref{sec:SNN_analysis} with \tablename~\ref{tab:discussion_on_macro_architecture}, the effect of the architecture components is consistently observed.
The use of the global average pooling layer and employing trainable spiking blocks or average pooling layers for down-sampling decreases the energy efficiency of SNNs, suggesting to exclude these design choices from SNN architectures.

\begin{table}[t]
\centering
\caption{Evaluation for different design choices on CIFAR10.}
\resizebox{\linewidth}{!}{
\addtolength{\tabcolsep}{-2pt}
\begin{tabular}{l|ccc|cc}
    \toprule
    Architecture & GAP & Normal & Down-sample & Acc.(\%) & Spikes \\
    \midrule
    SNN\_1 & \xmark & \texttt{SRB\_k3} & MaxPool &  87.54 & 146K  \\
    \midrule
    SNN\_2 & \cmark & \texttt{SRB\_k3} & MaxPool & 85.82 & 168K  \\
    \midrule
    SNN\_3 & \xmark & \texttt{SRB\_k3} & \texttt{SRB\_k3} & 89.18 & 221K  \\
    SNN\_4 & \xmark & \texttt{SRB\_k3} & AvgPool & 83.79 & 291K \\
    \bottomrule
\end{tabular}
\addtolength{\tabcolsep}{2pt}
}
\label{supp_tab:discussion_on_macro_architecture_SRB}
\end{table}

\subsection{Exploration for Energy-Efficient Candidate Blocks}\label{supp:SIB_analysis}

\begin{figure}[t]
    \centering
    \includegraphics[width=\linewidth]{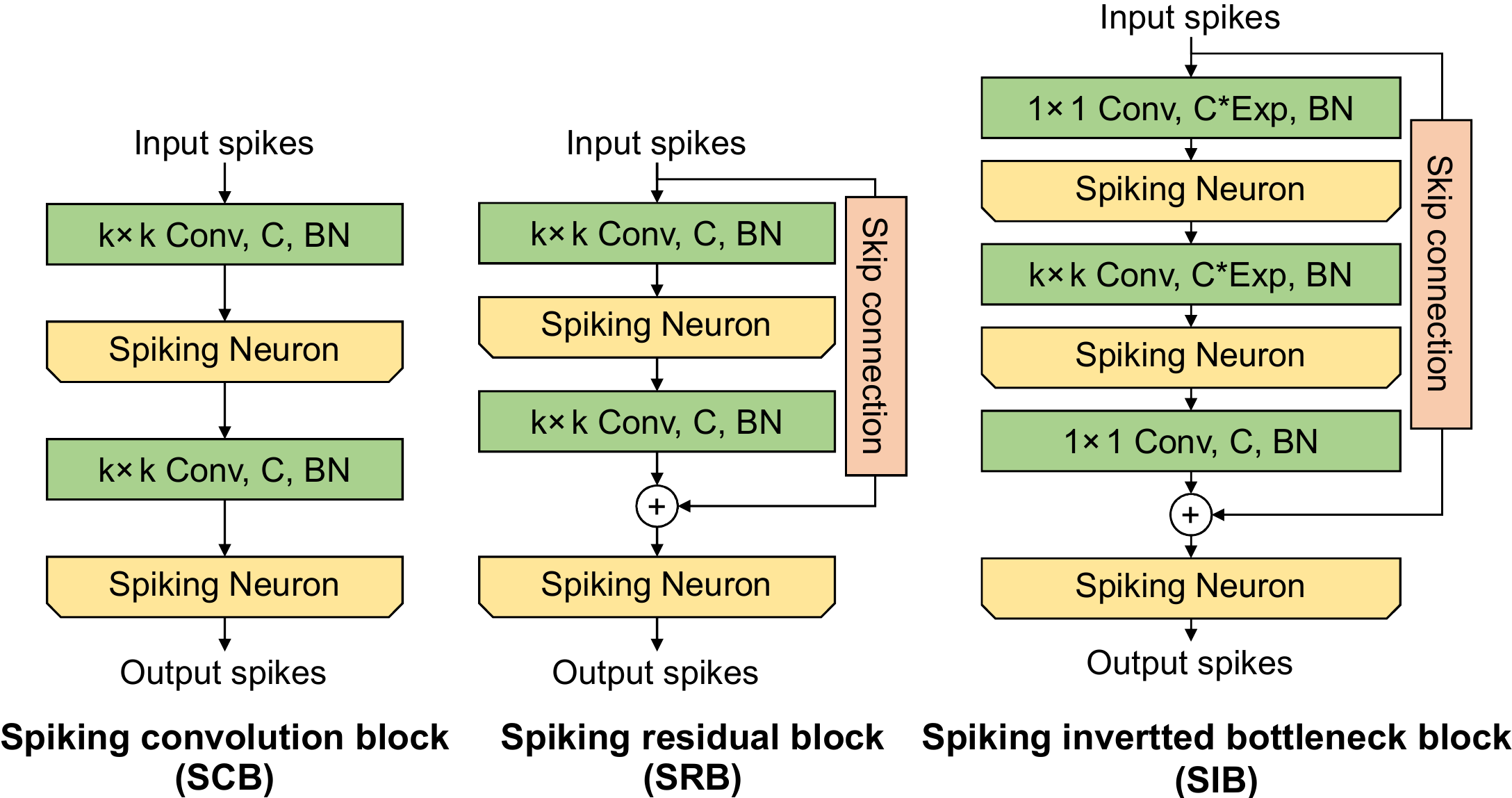}
    \vspace{-1em}
    \caption{Spiking blocks, where a convolution layer, channels, and batch normalization are denoted by Conv, C, and BN, respectively.} 
    \label{fig:spiking_blocks}
\end{figure}

In this study, we investigated which spiking blocks are suitable for designing energy-efficient SNNs.
In previous studies, building blocks based on SCB and SRB have been mainly used.
We newly introduced spiking inverted bottleneck block (SIB), inspired by the inverted bottleneck structure of MobileNetV2~\cite{sandler2018mobilenetv2}.
Three trainable spiking blocks that we standardized are depicted in \figurename~\ref{fig:spiking_blocks}.
The inverted bottleneck structure can reduce the number of parameters and FLOPs in ANNs and is thus widely used to search for ANNs suitable to mobile devices~\cite{cai2019proxylessnas,wu2019fbnet,tan2019mnasnet,cai2020ofanet}.
The hardware affinity of the inverted bottleneck structure can be in line with designing SNN architectures that are realized on neuromorphic chips. 
The design choices for SIB include kernel size $k$ and expansion ratio $e$; for simplicity, we denote SIB with $k=3$ and $e=3$ as \texttt{SIB\_k3\_e3}.

\begin{table}[t]
\centering
\caption{Evaluation for SNN\_1-based architectures, which consist of a single type of spiking block.}
\addtolength{\tabcolsep}{-2.5pt}
\begin{tabular}{l|ccc}
    \toprule
    Spiking block in SNN\_1 & Acc. (\%) & Spikes & Firing rates \\
    \midrule
    \texttt{SCB\_k3} & 86.93 & 154K & 0.18 \\
    \texttt{SRB\_k3} & 87.54 & 146K & 0.17\\
    \texttt{SIB\_k3\_e1} & 81.07 & 243K & 0.23 \\
    \texttt{SIB\_k3\_e3} & 88.45 & 374K & 0.17 \\
    \bottomrule
\end{tabular}
\addtolength{\tabcolsep}{2.5pt}
\label{tab:exploration_candidate_blocks}
\end{table}

\begin{figure}[t]
    \centering
    \includegraphics[width=\linewidth]{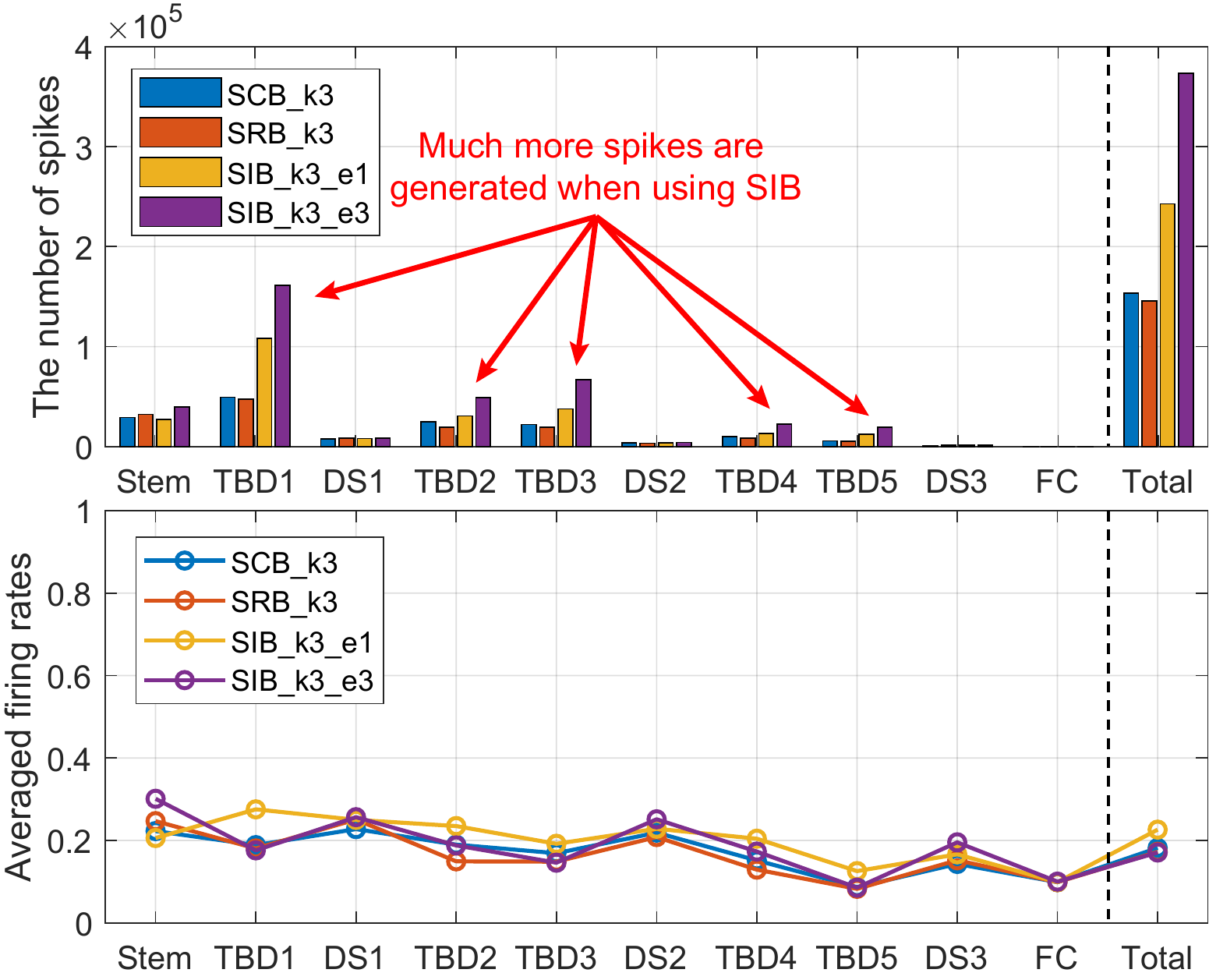}
    \vspace{-1em}
    \caption{The number of spikes generated by architectures whose TBD blocks are assigned by a single type of spiking block, and their firing rates averaged over test data and 8 timesteps.}
    \label{supp_fig:all_blocks}
\end{figure}

Three trainable spiking blocks, \ie, SCB, SRB, and SIB, are evaluated on SNN\_1.
\tablename~\ref{tab:exploration_candidate_blocks} and \figurename~\ref{supp_fig:all_blocks} provide the results when TBD blocks in SNN\_1 are assigned a single type of block with a kernel size of 3.
The SIB is significantly less desirable for energy-efficient SNNs than the SCB and SRB, even though the SNN with \texttt{SIB\_k3\_e3} improves the accuracy.
The SNNs with \texttt{SIB\_k3\_e1} and \texttt{SIB\_k3\_e3} generate approximately 1.6x and 2.4x more spikes than that with \texttt{SCB\_k3}, respectively.
Because the firing rate of each TBD block is similar across all three blocks (the second plot in \figurename~\ref{supp_fig:all_blocks}), the difference in the number of spikes can be associated with the change in the number of spiking neurons.

The number of neurons of each block is theoretically as follows: $2hwc$ for \texttt{SCB\_k3} (which is identical for \texttt{SRB\_k3}), $hw(2c_\mathrm{in} + c_\mathrm{out})$ for \texttt{SIB\_k3\_e1}, and $hw(6c_\mathrm{in} + c_\mathrm{out})$ for \texttt{SIB\_k3\_e3}, where $hw$ is a resolution of its input feature map, and $c_\mathrm{in}$ and $c_\mathrm{out}$ are the number of channels of its input and output feature maps, respectively.
Using the above equations, we obtain the number of spiking neurons of SNN\_1 with \texttt{SCB\_k3}, \texttt{SIB\_k3\_e1}, and \texttt{SIB\_k3\_e3}: approximately $6.4HWC$, $8.0HWC$, and $16.5HWC$, respectively, where $HW$ is a resolution of an image and $C$ is an initial channel of the SNN.
By applying the total firing rates (0.18, 0.23, and 0.17, respectively), we can calculate the approximated number of spikes as follows: $1.17HWC$, $1.84HWC$, and $2.81HWC$, respectively.
Hence, we can estimate that SNNs with \texttt{SIB\_k3\_e1} and \texttt{SIB\_k3\_e3} generate approximately 1.6x and 2.4x more spikes than that with \texttt{SCB\_k3}.
It is confirmed that the empirical results in \tablename~\ref{tab:exploration_candidate_blocks} and our theoretical calculation are consistent.
Therefore, to obtain energy-efficient SNNs, it is undesirable to include SIBs that have a large number of spiking neurons.

\noindent\textbf{Applying spike regularization}
Using a method to regularize the spike generation~\cite{pellegrini2021spike_regularization}, we further evaluate spiking blocks with the following spike regularization term: $\frac{\lambda_\mathrm{reg}}{KT} \sum_t \sum_k \phi_k [t]$, where $\phi$ is a spike, $K$ is the number of neurons, and $T$ is the timesteps.
This term is added to the training loss, and $\lambda_\mathrm{reg}$ controls the regularization strength.
The results are provided in \tablename~\ref{sub_tab:spike_regularization}.
Even with spike regularization, SIB is a less desirable choice than SCB and SRB in terms of both accuracy and spikes.

\begin{table}[h]
    \centering
    \caption{Combining \tablename~\ref{tab:exploration_candidate_blocks} with a regularization technique.}
    \resizebox{\columnwidth}{!}{
    \setlength{\tabcolsep}{2pt}
    \renewcommand{\arraystretch}{0.9}
    \begin{tabular}{ccccccccc}
        \toprule
        Spiking & \multicolumn{2}{c}{$\lambda_\mathrm{reg}=1$} & \multicolumn{2}{c}{$\lambda_\mathrm{reg}=0.1$} & \multicolumn{2}{c}{$\lambda_\mathrm{reg}=0.01$} & \multicolumn{2}{c}{$\lambda_\mathrm{reg}=0$}\\
        block & Acc. & Spikes & Acc. & Spikes & Acc. & Spikes & Acc. & Spikes \\
        \midrule
        \texttt{SCB\_k3} & 64.36 & 83K & 79.09 & 84K & 86.39 & 124K & 86.93 & 154K \\
        \texttt{SRB\_k3} & 72.76 & 49K & 83.25 & 70K & 86.59 & 109K & 87.54 & 146K \\
        \texttt{SIB\_k3\_e1} & 56.61 & 89K & 73.54 & 119K & 81.05 & 155K & 81.07 & 243K \\
        \texttt{SIB\_k3\_e3} & 74.71 & 136K & 84.59 & 186K & 87.61 & 249K & 88.45 & 374K \\
        \bottomrule
    \end{tabular}
    }
    \label{sub_tab:spike_regularization}
\end{table}

\section{Search Algorithm Implementation}\label{supp:search_algorithm}

\begin{figure}[t]
    \centering
    \includegraphics[width=\linewidth]{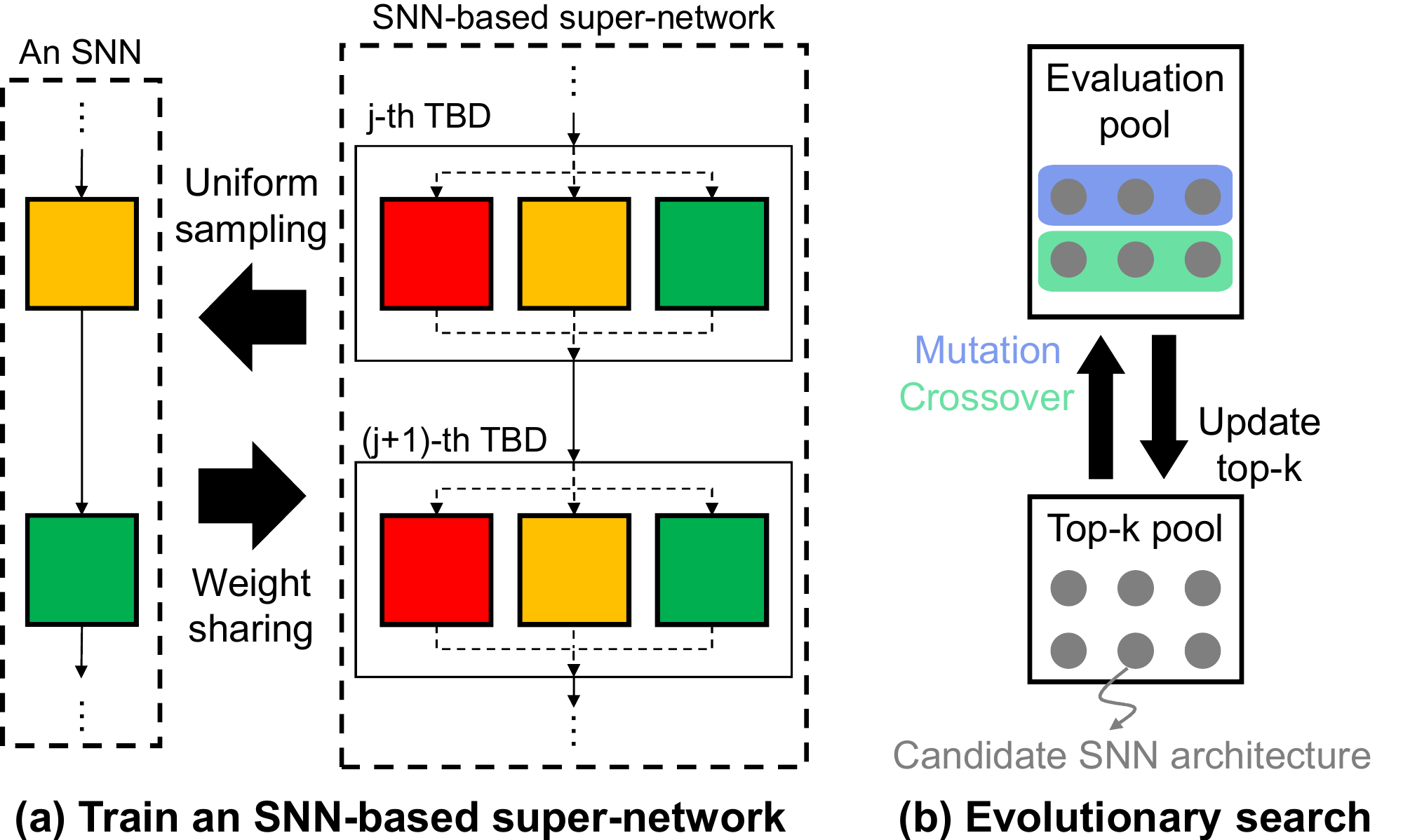}
    \vspace{-1em}
    \caption{Two consecutive searching procedures of AutoSNN. The colored blocks in (a) represent candidate spiking blocks.}
    \label{fig:overview_AutoSNN}
\end{figure}

\begin{algorithm}[tb]
\caption{Evolutionary search algorithm of AutoSNN}
\label{alg:search_algorithm}
\textbf{Input}: Trained super-network $S(W^\ast)$, validation data $D_\mathrm{val}$\\
\textbf{Parameter}: Fitness coefficient $\lambda$, max round $\mathcal{T}$, mutation ratio $\rho$, the number of architectures in evolutionary way $p_m$ and $p_c$, top-$k$ pool size $k$, evaluation pool size $p$\\
\textbf{Output}: SNN $A^\ast$ with the highest fitness

\begin{algorithmic}[1] 
    \STATE \textrm{P$_\textrm{eval}$} = \textrm{P$_\textrm{top}$} = $\emptyset$
    \FOR {r = 1 : $\mathcal{T}$ }
    \IF {r == 1}
        \STATE \textrm{P$_\textrm{eval}$} = RandomSample($p$)
    \ELSE
        \STATE P$_1$ = Mutation(\textrm{P$_\textrm{top}$}, $p_m$, $\rho$)
        \STATE P$_2$ = Crossover(\textrm{P$_\textrm{top}$}, $p_c$)
        \STATE \textrm{P$_\textrm{eval}$} = P$_1$ $\cup$ P$_2$
        \IF {Size(P$_\textrm{eval}$) $< p$}
            \STATE P$_3$ = RandomSample($p$ - Size(P$_\textrm{eval}$))
            \STATE \textrm{P$_\textrm{eval}$} = P$_\textrm{eval}$ $\cup$ P$_3$
        \ENDIF
    \ENDIF
    \STATE fitness values = Evaluate($S(W^\ast$), $D_\mathrm{val}$, \textrm{P$_\textrm{eval}$}, $\lambda$)
    \STATE \textrm{P$_\textrm{top}$} = UpdateTop$k$(\textrm{P$_\textrm{top}$}, \textrm{P$_\textrm{eval}$}, fitness values)
    \ENDFOR
    \STATE \textbf{return} Top-1 SNN architecture $A^\ast$ in \textrm{P$_\textrm{top}$}
\end{algorithmic}

\end{algorithm}

AutoSNN consists of two separate procedures, as illustrated in~\figurename~\ref{fig:overview_AutoSNN}: super-network training with a direct training method of SNNs and an evolutionary search algorithm with a spike-aware fitness.
Using the proposed spike-aware fitness, AutoSNN finds the architecture with the highest fitness value through an evolutionary search algorithm.
Throughout the search process, depicted by~\figurename~\ref{fig:overview_AutoSNN}(b), AutoSNN maintains two population pools: the top-$k$ population pool \textrm{P$_\textrm{top}$} with size $k$ and the temporary evaluation population pool \textrm{P$_\textrm{eval}$} with size $p$.
Along with Algorithm~\ref{alg:search_algorithm}, we provide a detailed explanation of AutoSNN as follows.

\textbf{Preparation (lines 3-13)}
First, \textrm{P$_\textrm{eval}$} is prepared with $p$ architectures, which are randomly sampled from the search space or generated through mutation and crossover.
We denote the number of architectures generated by mutation and crossover as $p_m$ and $p_c$, respectively. 
In the first round, all $p$ architectures are randomly sampled.
For mutation, a parent architecture $M$ is sampled from \textrm{P$_\textrm{top}$}, and each block in $M$ is stochastically mutated with a mutation ratio $\rho$.
For crossover, parent architectures $M_1$ and $M_2$ are sampled from \textrm{P$_\textrm{top}$}.
The offspring architecture is generated by stacking the first $X$ blocks in $M_1$ and the last $(5-X)$ blocks in $M_2$; the value of $X$ is randomly sampled from $\{1, 2, 3, 4\}$.
If an architecture that has already been evaluated is generated through mutation or crossover, it is not joined into the evaluation pool.
As the search proceeds, and the architectures in \textrm{P$_\textrm{top}$} start to remain unchanged, it may become difficult to obtain a new architecture through crossover.
When no new architectures are obtained through mutation or crossover, we fill \textrm{P$_\textrm{eval}$} by randomly sampling architectures.

\textbf{Evaluation and \textrm{P$_\textrm{top}$} Update (lines 14 and 15)}
All architectures in \textrm{P$_\textrm{eval}$} are evaluated based on their fitness values calculated using $D_\mathrm{val}$.
To update \textrm{P$_\textrm{top}$}, the top-$k$ architectures are selected from \textrm{P$_\textrm{eval}$} and \textrm{P$_\textrm{top}$} of the previous round, based on their fitness values.

After repeating the aforementioned processes for $\mathcal{T}$ iterations, AutoSNN obtains architecture $A^\ast$ with the highest fitness value from \textrm{P$_\textrm{top}$}.
For all experiments, we set the parameters used in Algorithm~\ref{alg:search_algorithm} as follows: $\mathcal{T} = 10$, $\rho = 0.2$, $p_m = 10$, $p_c = 10$, $k = 10$, and $p = 20$.

\section{Supplemental Results of AutoSNN}

\subsection{SNN Architectures Searched with Different $\lambda$}\label{supp:different_lambda}

\begin{figure}[t]
    \centering
    \includegraphics[width=\linewidth]{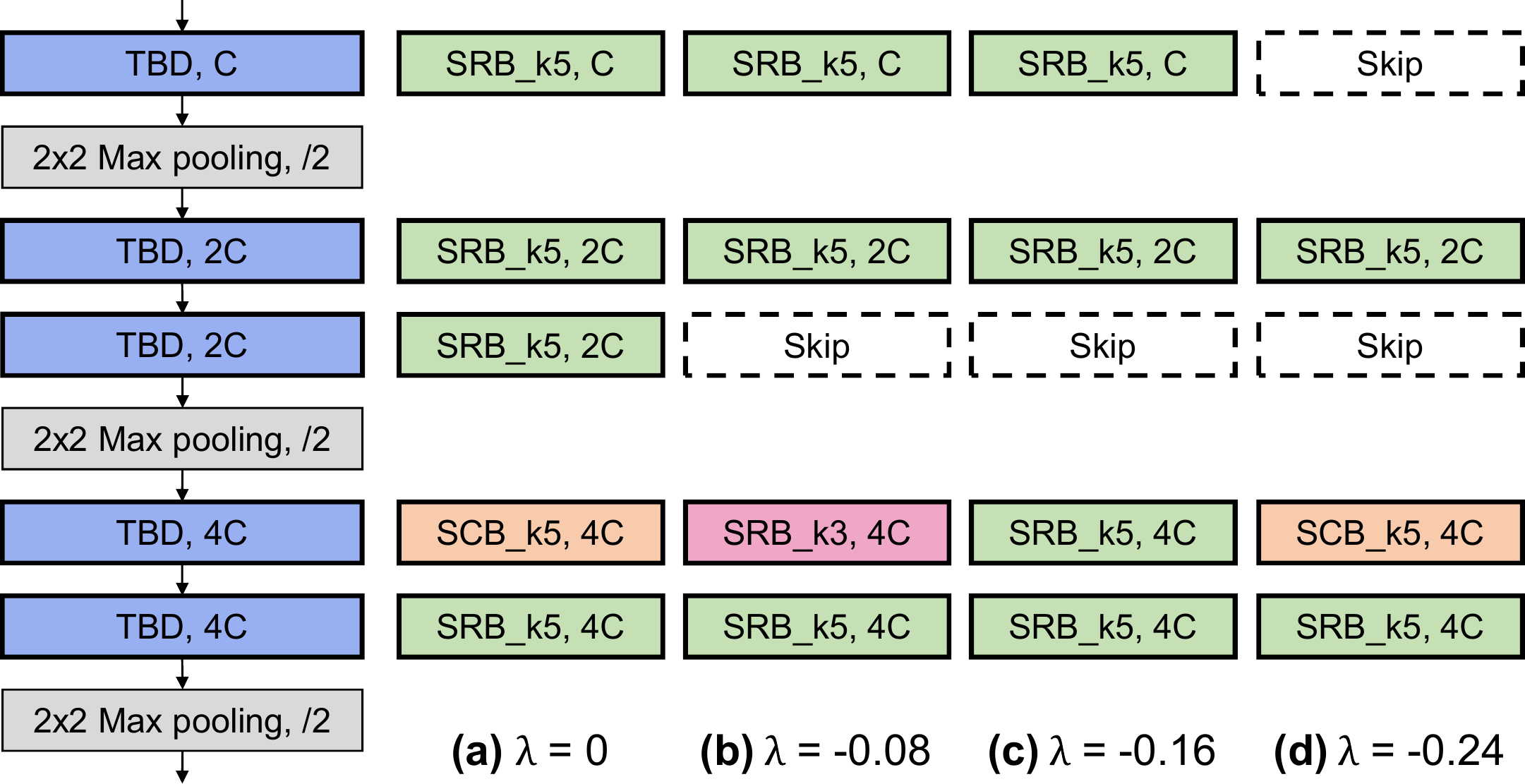}
    \vspace{-1em}
    \caption{SNN architectures searched by AutoSNN with different $\lambda$ on the proposed search space.}
    \label{supp_fig:searched_arch_lambda}
\end{figure}

SNN architectures searched by AutoSNN with different $\lambda$ of the fitness value are visualized in \figurename~\ref{supp_fig:searched_arch_lambda}.
For $\lambda=0$, indicating that only accuracy is considered during the search process, the searched architecture includes five spiking blocks with trainable parameters, and thus has the highest model complexity among the searched architectures.
For $\lambda=-0.08$ and $-0.16$, the third TBD block in the architectures is filled with \texttt{Skip}.
As shown in Table~\ref{tab:lambda_change}, these architectures experience a slight drop in accuracy while reducing approximately 20K spikes, compared to the architecture for $\lambda=0$.
In the architecture searched with $\lambda=-0.24$, both the first and third TBD blocks are filled with \texttt{Skip}.
The spikes generated by this architecture are much fewer than the other searched architectures, but the accuracy decreases by approximately 2\%p.
These searching results confirm that $\lambda$ functions according to our intent to adjust the accuracy-efficiency trade-off in the searched SNN.

Furthermore, we observe that the architectures searched on the proposed SNN search space prefer spiking blocks with a kernel size of 5 (\ie, \texttt{SCB\_k5} and \texttt{SRB\_k5}).
It would be also interesting to investigate architectural properties related to such preference of SNNs.

\subsection{Architecture Search without Spiking Neurons}

\begin{figure}[t]
    \centering
    \includegraphics[width=0.47\linewidth]{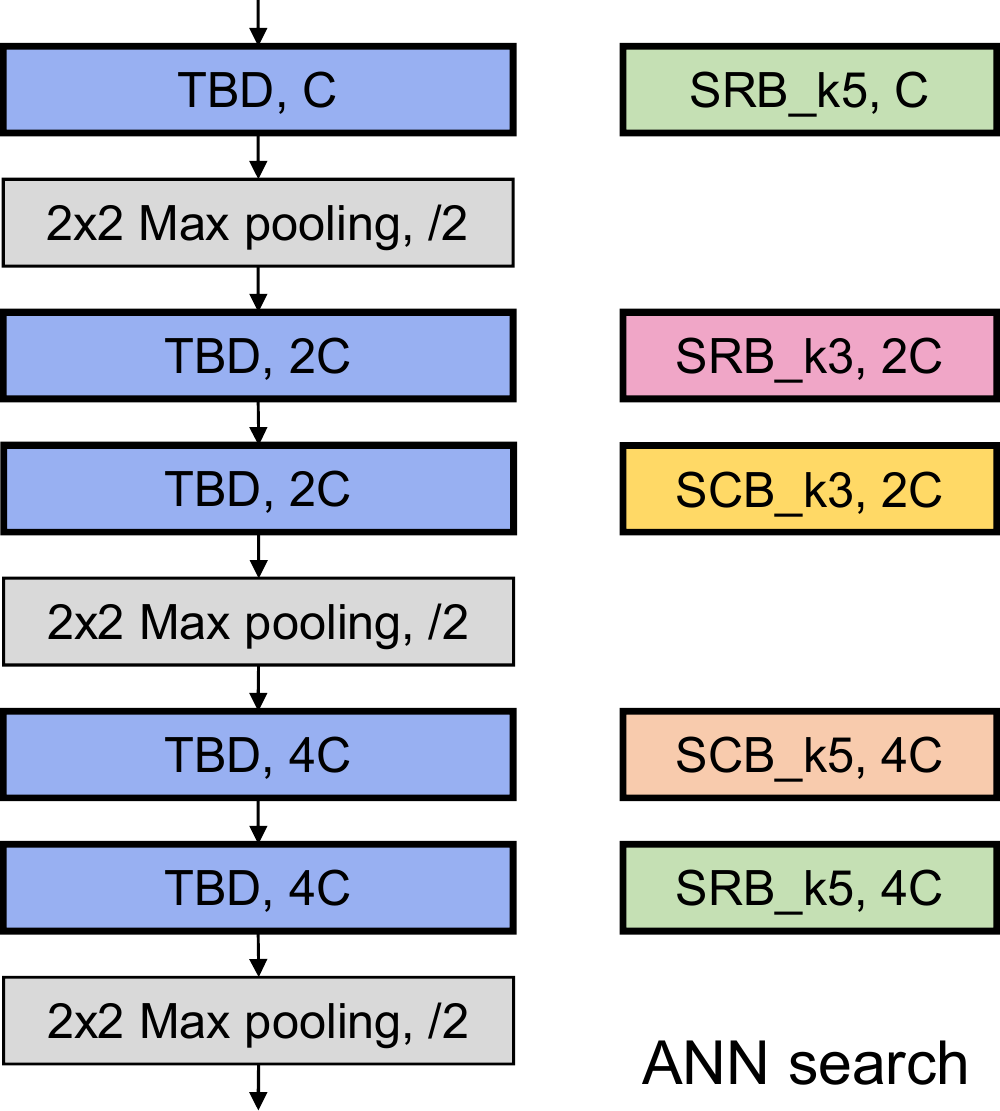}
    \vspace{-1em}
    \caption{An architecture which is searched from the ANN search space. The candidate blocks filling TBD blocks in the architecture are inverted into the corresponding spiking blocks.}
    \label{supp_fig:ANN_search}
\end{figure}

In Section~\ref{subsec:ANN_search}, we validate the importance of including spiking neurons in the search space.
\figurename~\ref{supp_fig:ANN_search} shows the architecture which is searched by our search algorithm on the search space without spiking neurons and then transformed into the SNN by adding spiking neurons to every selected blocks.

\subsection{Comparison with Recent SNN Techniques}

\begin{table}[t]
\centering
\caption{Comparison with Recent SNN Techniques: vs. manually modified SNNs (the first row group) and vs. SNNs searched by a NAS approach (the second row group)}
\resizebox{\columnwidth}{!}{
\begin{tabular}{cccc}
        \toprule
        Method & Acc.(\%) & Spikes & Timesteps \\
        \midrule
        VGG9+BNTT ($C=128$) & 90.5 & 131K & 25\\
        VGG16\_SBN+SALT ($C=64$) & 87.1 & 1500K & 40  \\
        AutoSNN ($C=16$) & 88.7 & 108K & 8  \\
        AutoSNN ($C=32$) & 91.3 & 176K & 8  \\
        \midrule
        SNASNet-Fw ($C=256$) & 93.6 & - & 8 \\
        SNASNet-Bw ($C=256$) & 94.1 & - & 8 \\
        AutoSNN ($C=128$) & 93.2 & 310K & 8 \\
        \bottomrule
    \end{tabular}
}
\label{sub_tab:comparison}
\end{table}

Here, in Table~\ref{sub_tab:comparison}, we compare AutoSNN with BNTT~\cite{kim2020BNTT}, SALT~\cite{kim2021SALT}, and SNASNet~\cite{kim2022SNASNet} as fairly as possible using their reported results.
In BNTT and SALT, architectures were manually modified; BNTT considers time variance of batch normalization (BN) layers in VGG9, and VGG16\_SBN used in SALT has a single BN layer before the last FC layer.
As described in Section~\ref{sec:related_work}, SNASNet is concurrent to our work in terms of utilizing NAS; but, SNASNet is based on totally different approach from AutoSNN, because SNASNet employed a NAS method that does not require any training process.
Compared to SALT, AutoSNN yielded higher accuracy and fewer spikes with less timesteps.
BNTT reduced the number of spikes but only at the cost of accuracy.
BNTT also uses a larger channel number and more timesteps than AutoSNN, both of which result in more energy consumption and longer latency.
Two SNASNet architectures with $C=256$ brought upon some increase in accuracy, but based on the results from the main paper, it is fair to expect AutoSNN with $C=256$ will also achieve comparable accuracy.

\section{Direct Training Framework for SNNs}\label{supp:direct_training}

In this section, we explain a direct training framework based on the supervised learning approach.
Assuming a classification task with $C$ classes, a loss function $L$ is defined as:
\begin{align}
    L(\textbf{o}, \textbf{y}) & = L(\frac{1}{T} \sum^{T-1}_{t=0} \textbf{$\phi$}^\mathrm{FC} [t], \textbf{y}),
\end{align}
where $T$ is the timesteps, $\textbf{y}$ is a target label, and $\textbf{o} \in \mathbb{R}^C $ is a predicted output, which is calculated by averaging the number of spikes generated by the last fully-connected (FC) layer over $T$.
In this study, we use the mean squared error (MSE) for $L$:
\begin{align}
     L = MSE(\textbf{o}, \textbf{y}) & = \frac{1}{C} \sum^{C-1}_{i=0} (\frac{1}{T} \sum^{T-1}_{t=0} \phi^\mathrm{FC}_i [t] - y_i)^2,
\end{align}
where $C$ is the number of classes.

\begin{figure}[t]
    \centering
    \includegraphics[width=\linewidth]{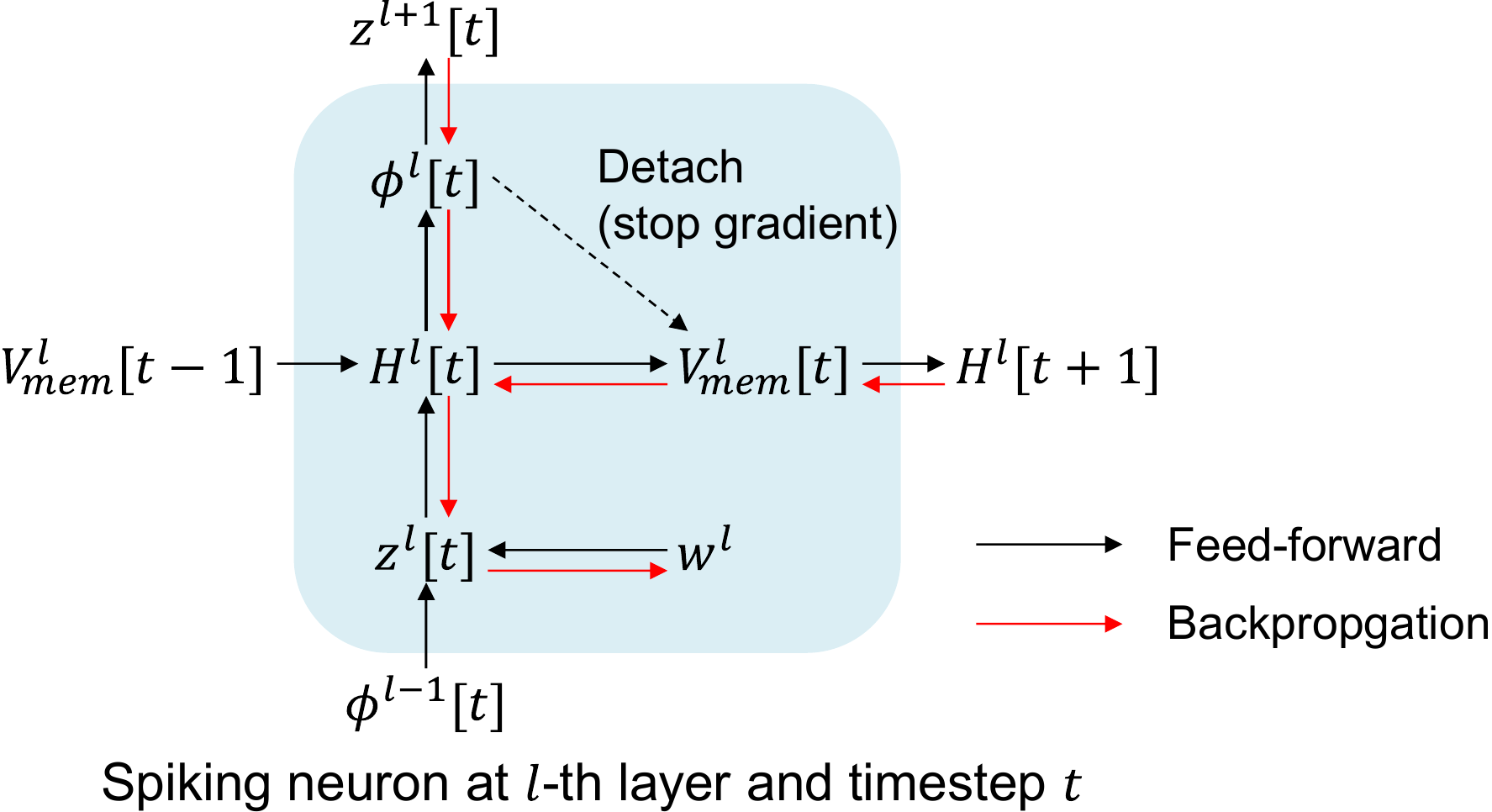}
    \vspace{-2em}
    \caption{Dynamics of a spiking neuron: feed-forward and backpropagation flows.}
    \vspace{-1em}
    \label{fig:spiking_neuron_backprop}
\end{figure}

As shown in \figurename~\ref{fig:spiking_neuron_backprop}, spikes in SNNs propagate through both spatial domain (from a lower layer to a higher layer) and temporal domain (from a previous timestep to a later timestep).
Hence, derivatives should be considered in the both perspectives.
We provide derivatives of components in a spiking neuron at the $l$-th layer and timestep $t$; these derivatives are highlighted by red lines in \figurename~\ref{fig:spiking_neuron_backprop}.
$w^l$ is shared across $T$ timesteps, and thus the derivative with respect to $w^l$ can be obtained according to the chain rule as follows:
\begin{align}\label{eq:L_w}
    \frac{\partial L}{\partial w^l} & = \sum^{T-1}_{t=0} \frac{\partial L}{\partial H^l [t]} \frac{\partial H^l [t]}{\partial z^l [t]} \frac{\partial z^l [t]}{\partial w^l}.
\end{align}
From Eq~\ref{eq:LIF_dynamics_discrete}, the second and third terms in the right-hand side of Eq.~\ref{eq:L_w} can be induced as follows:
\begin{align}
    \frac{\partial z^l [t]}{\partial w^l} & = \phi^{l-1}[t], ~~~~~~~~
    \frac{\partial H^l [t]}{\partial z^l [t]} = \frac{1}{\tau_\mathrm{decay}}.
\end{align}
By applying the chain rule to the first term, $\frac{\partial L}{\partial H^l [t]}$ is written as:
\begin{align}\label{eq:L_H}
    \frac{\partial L}{\partial H^l [t]} & = \frac{\partial L}{\partial \phi^l [t]} \frac{\partial \phi^l [t]}{\partial H^l [t]} + \frac{\partial L}{\partial V_\mathrm{mem}^l [t]} \frac{\partial V_\mathrm{mem}^l [t]}{\partial H^l [t]}.
\end{align}
The first two terms and the last two terms are related to the spatial and temporal domains, respectively.
In the spatial domain, $\frac{\partial L}{\partial \phi^l [t]}$ is obtained by using the spatial gradient back-propagated from the ($l+1$)-th layer as follows:
\begin{align}
    \frac{\partial L}{\partial \phi^l [t]} & = \frac{\partial L}{\partial z^{l+1} [t]} \frac{\partial z^{l+1} [t]}{\partial \phi^l [t]} 
    = \frac{\partial L}{\partial z^{l+1} [t]} w^{l+1}.
\end{align}
Note that SNNs have the non-differentiable property due to the spike firing $\Theta(x)$.
Approximation for the derivative of a spike, \ie, $\Theta'(x)$, is necessary to optimize SNNs by gradient-based training.
In this dissertation, we approximate $\Theta'(x) = \frac{1}{1+x^2}$ by employing the inverse tangent function for $\Theta (x) = \mathrm{arctan}(x)$.
Therefore, $\frac{\partial \phi^l [t]}{\partial H^l [t]}$ is calculated using the approximation and $\phi^l [t]$ in Eq.~\ref{eq:LIF_dynamics_discrete}:
\begin{align}
    \frac{\partial \phi^l [t]}{\partial H^l [t]} & = \Theta'(H^l [t] - V_\mathrm{th}).
\end{align}
The derivatives in the temporal domain of Eq.~\ref{eq:L_H} are also induced from Eq.~\ref{eq:LIF_dynamics_discrete} as follows:
\begin{align}
    \frac{\partial L}{\partial V_\mathrm{mem}^l [t]} & = \frac{\partial L}{\partial H^l [t+1]} \frac{\partial H^l [t+1]}{\partial V_\mathrm{mem}^l [t]} \\
        & = \frac{\partial L}{\partial H^l [t+1]} (1 - \frac{1}{\tau_\mathrm{decay}}), \\
    \frac{\partial V_\mathrm{mem}^l [t]}{\partial H^l [t]} & = 1 - \phi^l [t] - H^l [t] \frac{\partial \phi^l [t]}{\partial H^l [t]}.
\end{align}


\end{document}